\documentclass[lettersize,journal]{IEEEtran}
\usepackage{cite}
\usepackage{graphicx}
\usepackage{amsmath}
\usepackage{array}
\usepackage{fixltx2e}
\usepackage{multirow}
\usepackage{url}
\usepackage{epsfig}
\usepackage{dblfloatfix}
\usepackage{graphicx}
\usepackage{amssymb}
\usepackage{bm}
\usepackage{xcolor}
\usepackage{float}
\usepackage{caption}
\usepackage{subcaption}
\usepackage{algorithm}
\usepackage{algpseudocode}
\usepackage{mathrsfs}
\usepackage[misc]{ifsym}
\usepackage[english]{babel}
\usepackage{xspace}
\usepackage{ragged2e}
\usepackage{gensymb}
\usepackage{pifont}
\usepackage{amsthm}
\usepackage{amsfonts}
\usepackage{makecell}
\usepackage{setspace}
\usepackage{colortbl}
\usepackage{booktabs}
\usepackage{footnote}
\usepackage{animate}
\usepackage{bbding}
\usepackage{arydshln}

\usepackage[pagebackref,breaklinks,colorlinks]{hyperref}
\let\MYoriglatexcaption\caption
\renewcommand{\caption}[2][\relax]{\MYoriglatexcaption[#2]{#2}}

\makeatletter
\DeclareRobustCommand\onedot{\futurelet\@let@token\@onedot}
\def\@onedot{\ifx\@let@token.\else.\null\fi\xspace}

\makeatother

\definecolor{BestRed}{RGB}{192,0,0}      
\definecolor{SecondBlue}{RGB}{0,112,192} 
\newcommand{\best}[1]{\textbf{\textcolor{BestRed}{#1}}}
\newcommand{\second}[1]{\textcolor{SecondBlue}{#1}}

\definecolor{BlueGreen}{rgb}{0.0, 0.5, 0.5}  
\definecolor{RedOrange}{rgb}{1.0, 0.27, 0.0}  

\makeatletter
\newcommand{\thickhline}{%
    \noalign {\ifnum 0=`}\fi \hrule height 1pt
    \futurelet \reserved@a \@xhline
}
\makeatother
\begin{document}

\title{Splats in Splats++: Robust and Generalizable 3D Gaussian Splatting Steganography}

\author{Yijia~Guo$^{\dagger}$,~Wenkai~Huang$^{\dagger}$,~Tong~Hu$^{\dagger}$,~Gaolei~Li\ding{41},~Yang~Li,~Yuxin~Hong,~Liwen~Hu,~Xitong~Ling,~Jianhua~Li\ding{41}, \IEEEmembership{Senior~Member,~IEEE}, Shengbo~Chen,~Tiejun~Huang, \IEEEmembership{Senior~Member,~IEEE}, and~Lei~Ma\ding{41}
  \IEEEcompsocitemizethanks{
    \IEEEcompsocthanksitem Yijia~Guo,~Liwen~Hu~and~Tiejun~Huang are with the State Key Laboratory of Multimedia Information Processing and National Engineering Research Center of Visual Technology, School of Computer Science, Peking University, Beijing, 100871, China.
    \IEEEcompsocthanksitem Wenkai~Huang, Gaolei Li, and Jianhua Li are with Shanghai Key Laboratory of Integrated Administration Technologies for Information Security, School of Computer Science, Shanghai Jiao Tong University, Shanghai 200240, China.
    \IEEEcompsocthanksitem Tong~Hu,~and~Lei~Ma are with National Biomedical Imaging Center, College of Future Technology, Peking University, Beijing, 100871, China.
    \IEEEcompsocthanksitem 
    Yuxing~Hong~and Shengbo~Chen are with the School of Software, Nanchang University, Nanchang, Jiangxi 330031, China.
    \IEEEcompsocthanksitem 
    Xitong Ling is with the Institute of Biopharmaceutical and Health Engineering, Tsinghua Shenzhen International Graduate School, Tsinghua
    University, Shenzhen 518055, China. 
    \IEEEcompsocthanksitem \ding{41} Corresponding author: Lei Ma (lei.ma@pku.edu.cn), Gaolei Li  (gaolei\_li@sjtu.edu.cn), Jianhua Li (lijh888@sjtu.edu.cn).
     \IEEEcompsocthanksitem $^{\dagger}$ These authors contributed equally.
  }
}

\markboth{Under Review}{}
\IEEEtitleabstractindextext{
  \begin{abstract}

3D Gaussian Splatting (3DGS) has recently redefined the paradigm of 3D reconstruction, striking an unprecedented balance between visual fidelity and computational efficiency. As its adoption proliferates, safeguarding the copyright of explicit 3DGS assets has become paramount. However, existing invisible message embedding frameworks struggle to reconcile secure and high-capacity data embedding with intrinsic asset utility, often disrupting the native rendering pipeline or exhibiting vulnerability to structural perturbations.
In this work, we present \textbf{\textit{Splats in Splats++}}, a unified and pipeline-agnostic steganography framework that seamlessly embeds high-capacity 3D/4D content directly within the native 3DGS representation. Grounded in a principled analysis of the frequency distribution of Spherical Harmonics (SH), we propose an importance-graded SH coefficient encryption scheme that achieves imperceptible embedding without compromising the original expressive power. To fundamentally resolve the geometric ambiguities that lead to message leakage, we introduce a \textbf{Hash-Grid Guided Opacity Mapping} mechanism. Coupled with a novel \textbf{Gradient-Gated Opacity Consistency Loss}, our formulation enforces a stringent spatial-attribute coupling between the original and hidden scenes, effectively projecting the discrete attribute mapping into a continuous, attack-resilient latent manifold.
Extensive experiments demonstrate that our method substantially outperforms existing approaches, achieving up to \textbf{6.28 db} higher message fidelity, \textbf{3$\times$} faster rendering, and exceptional robustness against aggressive 3D-targeted structural attacks (e.g., GSPure). Furthermore, our framework exhibits remarkable versatility, generalizing seamlessly to 2D image embedding, 4D dynamic scene steganography, and diverse downstream tasks.

\end{abstract}
}

\maketitle
\IEEEdisplaynontitleabstractindextext
\IEEEpeerreviewmaketitle
\section{Introduction}
\label{sec:intro}

Embedding invisible information within public carriers is a fundamental challenge in computing, centered on the balance between perceptual transparency and statistical integrity \cite{baluja2017hiding, tancik2020stegastamp, zhang2019steganogan}. In the evolving digital landscape, such mechanisms are critical for safeguarding the intellectual property of high-value digital assets \cite{uchida2017embedding, xue2021intellectual}. As a dominant paradigm for 3D digital asset representation, 3D Gaussian Splatting (3DGS) \cite{3dgs} has been widely adopted in applications such as e-commerce visualization \cite{tang2024lgm}, extended reality (XR) \cite{lei2024gart, xiang2024flashavatar}, film production \cite{luiten2024dynamic, yang2024deformable}, robotics \cite{matsuki2024gaussian, keetha2024splatam}, and 3D generative modeling \cite{tangdreamgaussian, chung2025luciddreamer}. As these models increasingly constitute high-value and computationally intensive assets, their explicit and easily transferable structure exposes them to substantial intellectual property risks. This tension between asset value and structural vulnerability underscores the necessity of developing principled and robust information embedding mechanisms tailored to the 3DGS framework. Beyond asset protection, such a paradigm establishes a foundational pathway toward automated provenance tracking and copyright authentication in next-generation 3D digital content ecosystems.


In general, analogous to invisible information embedding in 2D representations and Neural Radiance Fields (NeRF) \cite{li2023steganerf}, an embedding framework tailored for 3DGS should satisfy three fundamental standards: \textbf{a) Usability,}
which ensures that the embedding process does not impair the primary utility of the original 3DGS and applicability to downstream tasks.
\textbf{b) Robustness,}
which guarantees the stability of the embedded information under common perturbations, processing operations, and potential removal attempts, thereby sustaining the persistence and reliability of the concealed content.
\textbf{c) Capacity,}
which enables the encoding of sufficiently rich hidden information and facilitates high-density integration that scales with the representational capability of the 3DGS carrier without degrading visual quality. These criteria are not independent in explicit radiance representations: 
incorporating high-capacity 3D information
inevitably alters geometric consistency, making robust embedding fundamentally a representation coupling problem rather than a signal hiding problem. Existing methods typically satisfy only a subset of the aforementioned criteria instead of providing a unified solution. 
For example, previous 3DGS steganography approaches \cite{zhang2024gshider,zhang2025securegs} degrade the functionality of the original representation, while 
bitstream-based watermarking schemes \cite{huang2024gaussianmarker,chen2025guardsplat,li2025gs} for 3DGS provide only limited embedding capacity. 
In addition, their robustness is largely confined to simple perturbations. \textbf{The field still lacks a principled and unified framework capable of jointly optimizing embedding capacity, robustness, and preservation of carrier usability.}


\begin{figure*}
  \centering
  \includegraphics[width=1\linewidth]{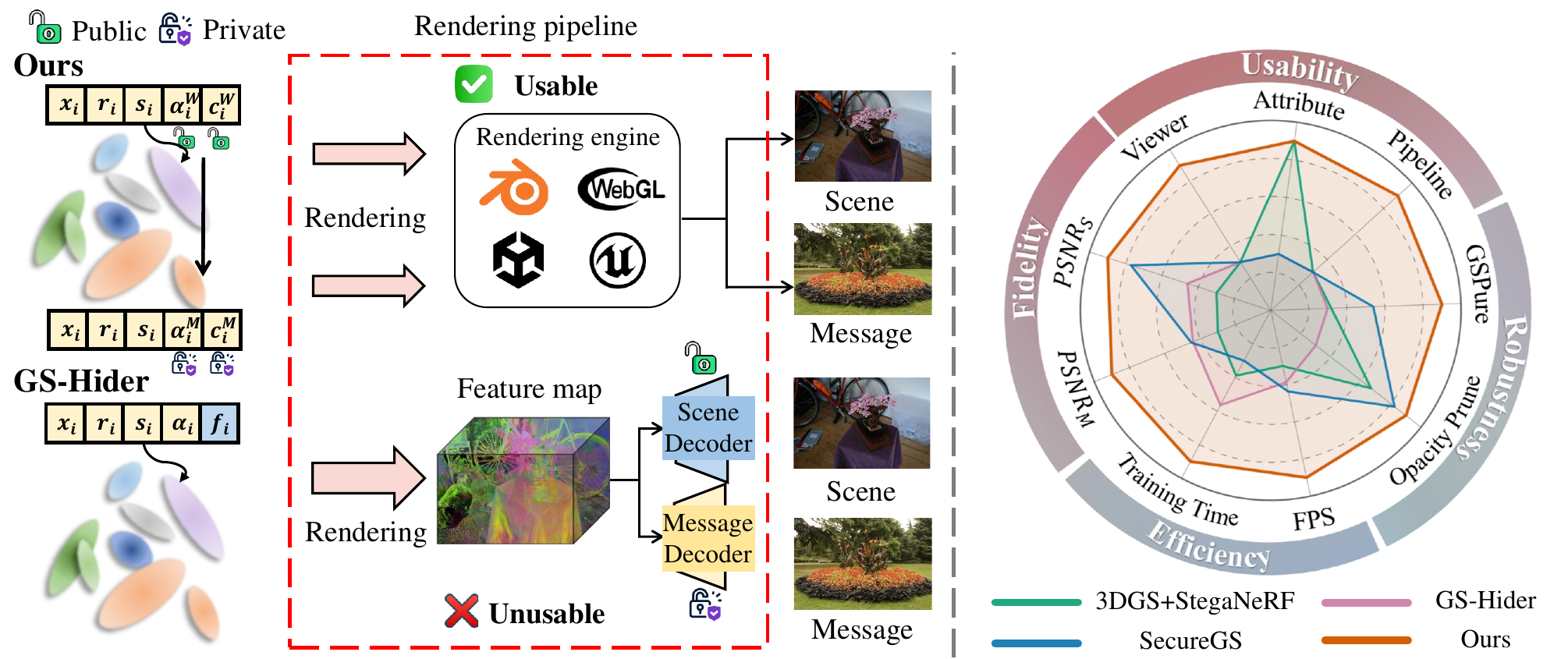}
  \captionof{figure}{\textbf{Left:} GS-Hider and our method's rendering pipeline. GS-Hider \cite{zhang2024gshider} employs a coupled feature field and neural decoders to render the original and hidden scenes simultaneously, affecting user’s conventional usage. We retain the vanilla 3DGS pipeline to preserve user experience. \textbf{Right:} Comparison of different 3DGS steganography methods. Existing works all have shortcomings in terms of robustness, fidelity, efficiency, and usability. Our method can maximize the fidelity of the original scene while ensuring the rendering speed and usability of 3DGS, as well as the security of the steganography. }
  \label{fig1}
\end{figure*}

In our previous work, Splats in Splats \cite{guo2024splats}, we leveraged the intrinsic properties of 3DGS to establish the first framework capable of embedding 3D content into vanilla 3DGS while fully preserving its original attributes, thereby addressing the capacity–usability-fidelity trade-off. However, the proposed method remains vulnerable to adversarial manipulations, and its generalizability across diverse scenarios has not yet been systematically validated. This reveals a deeper limitation: embedding that relies solely on attribute preservation fails when structural operations perturb the geometry–appearance correspondence inherent to 3DGS. Motivated by these limitations, we present \textbf{Splats in Splats++}, a powerful and general 3DGS steganography framework designed to achieve high-capacity, efficient, high-fidelity, and robust message embedding while preserving the practicality of vanilla 3DGS. 
To enable this capability, we \textbf{first} present a principled analysis of spherical harmonics (SH) in 3DGS, which underpins our framework. Exploiting the hierarchical importance of SH coefficients, we develop an importance-graded encryption and decryption scheme that selectively embeds hidden SH information into the original representation. This design achieves imperceptible integration of concealed 3D content while preserving the expressive power and downstream usability of the original 3DGS.
\textbf{Second}, to prevent message leakage caused by geometric ambiguities and enhance the geometric fidelity of the recovered hidden scene, we design a dedicated opacity mapping mechanism for the hidden scene. Specifically, a multi-resolution hash grid maps Gaussian coordinates into latent spatial descriptors, which are concatenated with opacity attributes and decoded by an MLP to predict the implicit opacity manifold. This design maintains high geometric fidelity for both the original and embedded scenes, while compactly encoding the key within a lightweight MLP.
\textbf{Third}, we introduce an opacity-gradient-driven gating mechanism and a consistency loss for steganographic training. By leveraging the spatial distribution of opacity gradients, this mechanism effectively enforces a stringent geometric coupling. Consequently, the resulting representations exhibit superior integration and heightened resilience against both structural perturbations and specialized adversarial attacks targeting 3DGS assets.  The contributions of this
work can be summarized as follows:

 
\begin{itemize}
\item[$\bullet$] 
 We propose Splats in Splats++, a robust and generalizable steganography framework that embeds 3D content in 3DGS itself without modifying any attributes of the vanilla 3DGS.  To the best of our knowledge, our method is also the first unified steganography scheme applicable to both the vanilla 3DGS architecture and its advanced extensions. 
\item[$\bullet$]
Building on a deep insight into the different order importance of SH coefficients, we propose an importance-graded coefficient encryption scheme that enables imperceptible embedding of hidden 3D content while preserving the inherent properties.
\item[$\bullet$]
We design a dedicated implicit opacity mapping mechanism, which leverages multi-resolution hash grid encoding and a lightweight MLP to achieve geometrically faithful embedding.
\item[$\bullet$] 
We employ an adaptive opacity control strategy to produce a tightly fused original–hidden scene representation that preserves visual fidelity while enhancing robustness to structural perturbations and 3D adversarial attacks.
\end{itemize}
Extensive experiments across multiple datasets and tasks demonstrate that Splats in Splats++ not only achieves superior visual fidelity but also reduces computational cost and strengthens robustness, highlighting its effectiveness across multiple dimensions of performance, as shown in Fig.~\ref{fig1}. Its exceptional versatility is confirmed by successful application to advanced scenarios, including 2D-3D and 3D-4D steganography. Our protected assets also ensure seamless integration into all standard rendering pipelines and downstream tasks.

Compared with our previous conference paper~\cite{guo2024splats}, in the journal version, Splats in Splats++ has the following new features: \textbf{a)} A opacity-gradient–based gating mechanism to enhance security and robustness against diverse 3D-targeted attacks.  \textbf{b)} A hash-grid guided opacity-mapping strategy to better align with the updated training scheme, achieving improved visual fidelity while reducing the key size. \textbf{c)} An extension to 4DGS steganography, demonstrating applicability to dynamic 3D assets.  \textbf{d)} Extensive supplementary experiments, including additional datasets, robustness analyses, downstream task evaluations, and 4D steganography results, confirming the effectiveness and practical viability of the proposed approach. These features enable Splats in Splats++ to achieve robust, generalizable, and extendable invisible information embedding for 3DGS.

\section{Related Works}
\label{sec:formatting}
\subsection{Radiance Field Reconstruction}
Neural Radiance Fields (NeRF) \cite{nerf} revolutionized 3D reconstruction by recovering volumetric radiance fields for bounded scenes. Subsequent advancements have extended NeRF to dynamic and non-rigid environments \cite{park2021nerfies, pumarola2021dnerf, li2021SceneFlownerf, yan2023nerfds}, enhanced rendering fidelity \cite{barron2021mipnerf, barron2022mipnerf360}, and improved robustness against severe motion blur \cite{qi2023e2nerf}.

More recently, 3DGS \cite{3dgs} has emerged as a formidable alternative, significantly accelerating training and rendering speeds through a discrete, point-based representation. This paradigm shift has catalyzed considerable research interest in optimizing its rendering quality \cite{yu2024mip}, computational efficiency \cite{fan2024instantsplat, chen2024mvsplat}, and storage requirements \cite{navaneet2024compgs}. 
The versatility of 3DGS is evidenced by its extension to diverse scenarios, including dynamic spatial-temporal modeling \cite{wu20234d, li2024st}, large-scale urban environments \cite{liu2024citygaussian, guo2025fly, meuleman2025fly}, and high-speed egomotion capture \cite{xiong2024event3dgs, yu2024evagaussians, guo2024spikegs}. Furthermore, researchers have successfully relaxed constraints regarding camera poses \cite{muller2022instant} and image sharpness \cite{chen2024deblur}, while expanding its utility into inverse rendering \cite{guo2024prtgs, liang2023gsir, gao2023relightable}, 3D content generation \cite{yi2023gaussiandreamer}, and interactive editing \cite{chen2023gaussianeditor}. As 3DGS increasingly becomes a mainstream format for 3D assets, the imperative for robust copyright management and provenance verification has emerged as a critical and urgent challenge.

\begin{figure*}[t]
  \centering
  \includegraphics[width=1\linewidth]{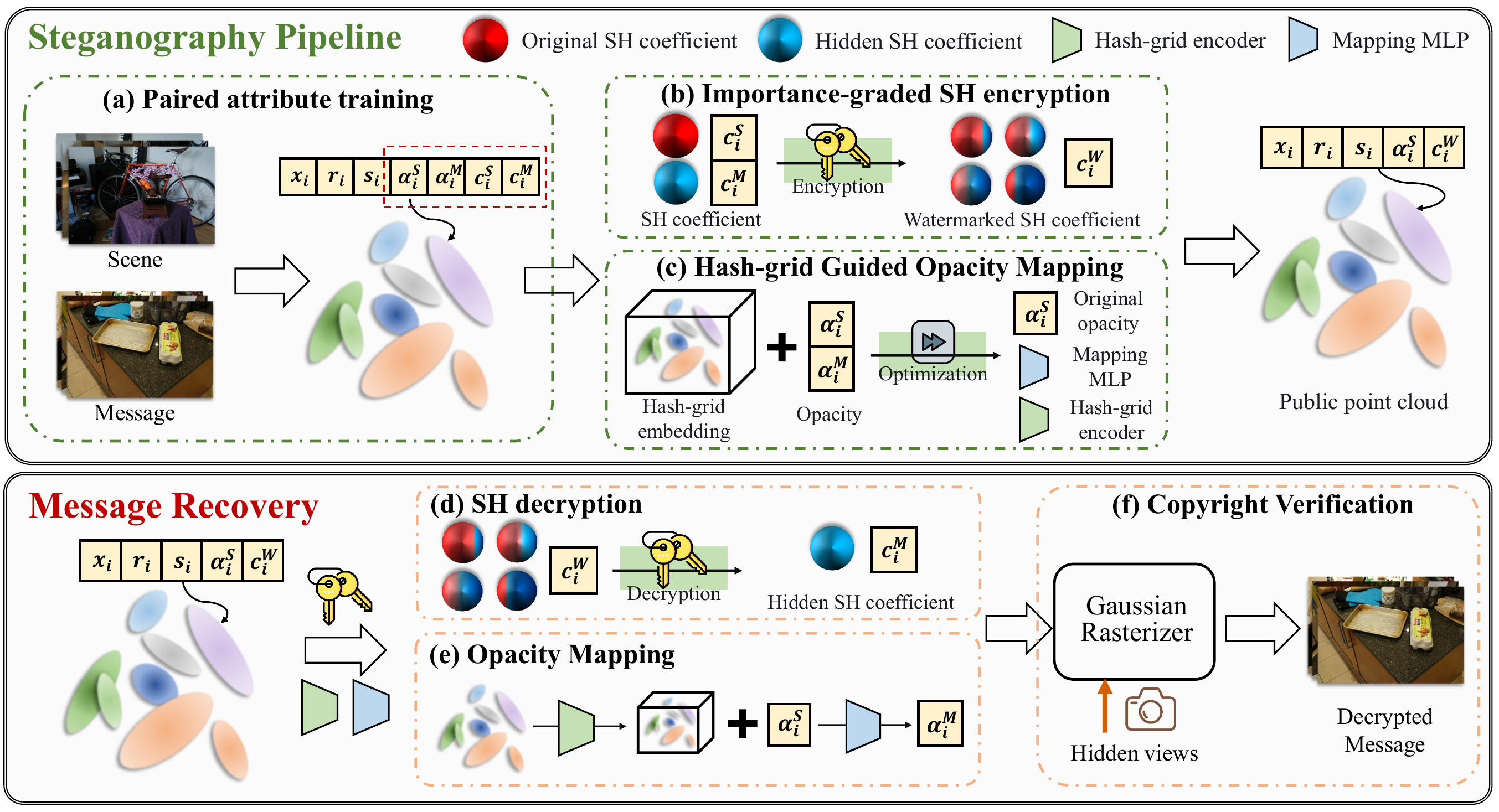}
  \caption{
  \textbf{Top: Steganography Pipeline. } We utilize the original and hidden views to train two sets of SH coefficients and opacity, while ensuring that both sets share the same Gaussian primitive locations. Training is supervised by rendering losses from the  original and hidden scenes, supplemented by an opacity gradient-based gating mechanism that facilitates the seamless fusion of both scene components. Following the training procedure, we perform steganography by employing importance-graded SH coefficient encryption (Sec.~\ref{Sec:Importance-graded SH Coefficient Encryption}) alongside hash-grid guided opacity mapping (Sec.~\ref{Sec:Hash-Grid Guided Opacity Mapping}). \textbf{Bottom: Message Recovery Pipeline. } The extraction process treats the hidden message recovery as a deterministic reconstruction from the fused 3DGS asset using a private key. Specifically, hidden SH coefficients are retrieved through bit-domain decryption of the public coefficients, while the hidden opacity field is reconstructed via neural field inference using the trained Hash-Grid guided mapping network.
  }
  \label{fig2}
\end{figure*}
\subsection{2D Steganography}
Learning-based 2D steganography methods are typically formulated as image-level post-processing techniques, where secret information is embedded into already generated images via neural transformations~\cite{baluja2017deep,zhu2018hidden,zhang2019udh,liu2019edn,ahmadi2020redmark}. In particular, HiDDeN~\cite{zhu2018hidden} demonstrates the feasibility of imperceptible yet decodable image modifications using deep networks, while ReDMark~\cite{ahmadi2020redmark} exploits residual representations to improve visual fidelity and robustness against common image processing operations. Overall, these approaches treat steganography as an external image manipulation procedure, largely independent of the image generation process itself.

With the emergence of diffusion-based generative models, recent studies have shifted toward embedding watermarks directly into the image synthesis process~\cite{fernandez2023stable,asnani2024promark,min2024watermark,yang2024gaussian,wen2024tree}. Stable Signature~\cite{fernandez2023stable} couples watermark information with latent diffusion dynamics, enabling generated images to inherently carry identifiable signatures. ProMark~\cite{asnani2024promark} further introduces proactive watermark injection during diffusion sampling to support causal attribution. Beyond single-owner scenarios, WaDiff~\cite{min2024watermark} incorporates watermark identifiers as conditioning variables, enabling multi-user distinguishable watermarking within a unified generative framework. More recently, Gaussian Shading~\cite{yang2024gaussian} proposes a theoretically grounded approach that achieves provable performance-lossless watermarking for diffusion models. Despite their effectiveness, these methods are inherently designed for 2D diffusion pipelines and do not readily generalize to explicit 3D representations.

\subsection{3D Steganography}
3D steganography techniques place greater emphasis on the unique characteristics of 3D representations rather than 2D images \cite{baluja2019hiding, xu2022robust}, videos \cite{mou2023large, luo2023dvmark} or large-scale generative models \cite{fernandez2023stable, wen2024tree}. Existing 3D steganography techniques primarily focus on 3D representations like meshes or point clouds, employing domain transformations \cite{li2023zero, kallel20233d} or homomorphic encryption \cite{van20233d} for information hiding. 

Recently, steganography techniques on implicit 3D representations like NeRF have gained prominence. StegaNeRF \cite{li2023steganerf} embeds secret images into 3D scenes by fine-tuning NeRF's weights while preserving the original visual quality. WaterRF \cite{jang2024waterf} introduces a steganography method for both implicit and explicit NeRF representations which embeds binary messages using discrete wavelet transform, achieving great performance in capacity, invisibility, and robustness. When considering 3DGS, a novel and efficient 3D representation, GS-Hider \cite{zhang2024gshider} utilized a scene decoder and a message decoder to disentangle the original scene from the hidden message, allowing for steganography within 3D scenes. However, GS-Hider not only caused a degradation in both rendering speed and visual quality but also damaged the vanilla 3DGS architecture, resulting in limited effectiveness in practical deployment. SecureGS \cite{zhang2025securegs}, which is built on Scaffold-GS \cite{scaffoldgs}, faces the same problems. So there is an urgent need for 3DGS steganography research that ensures copyright protection while maintaining the efficiency and usability of the vanilla 3DGS framework.

\section{Preliminary: 3D Gaussian Splatting}

The 3D Gaussian Splatting (3DGS) framework represents a 3D scene through a collection of anisotropic Gaussian primitives. Each primitive $i$ is spatially defined by a centric coordinate $\mathbf{x}_i \in \mathbb{R}^3$ and a 3D covariance matrix $\Sigma \in \mathbb{R}^{3 \times 3}$. The influence of a Gaussian primitive at position $\mathbf{x}$ is formulated as:
\begin{equation}
G(\mathbf{x}) = \exp \left( -\frac{1}{2} (\mathbf{x} - \mathbf{x}_i)^\top \Sigma^{-1} (\mathbf{x} - \mathbf{x}_i) \right).
\end{equation}
To maintain the physical validity of the covariance matrix during optimization, $\Sigma$ is typically decomposed into a rotation matrix $\mathbf{R}$ (represented by a normalized quaternion $\mathbf{q}_i$) and a scaling matrix $\mathbf{S} = \text{diag}(\mathbf{s}_i)$:
\begin{equation}
\Sigma = \mathbf{R} \mathbf{S} \mathbf{S}^\top \mathbf{R}^\top.
\end{equation}

For rendering, 3DGS utilizes a tile-based rasterizer to project 3D Gaussians onto a 2D camera plane. The projected 2D covariance matrix $\Sigma'$ is derived through the viewing transformation $\mathbf{W}$ and the Jacobian $\mathbf{J}$ of the affine approximation:
\begin{equation}
\Sigma' = \mathbf{J} \mathbf{W} \Sigma \mathbf{W}^\top \mathbf{J}^\top.
\end{equation}
The final color $\mathbf{C}$ of a pixel is synthesized by accumulating $N$ ordered primitives overlapping the pixel via alpha-blending:
\begin{equation}
\mathbf{C} = \sum_{i=1}^{N} \mathbf{c}_i \delta_i \prod_{j=1}^{i-1} (1 - \delta_j), \quad \delta_i = \alpha_i \cdot G'(\mathbf{x}'),
\label{eq:rendering}
\end{equation}
where $\delta_i$ denotes the effective opacity at the pixel coordinate $\mathbf{x}'$, $\alpha_i$ is the learned per-primitive opacity, and $\mathbf{c}_i$ is the view-dependent color represented by Spherical Harmonic (SH) coefficients.
In summary, a 3DGS model is essentially an attribute ensemble $\mathcal{A} = \{ \mathbf{x}_i, \mathbf{q}_i, \mathbf{s}_i, \alpha_i, \mathbf{c}_i \}_{i=1}^N$. Within this multidimensional representation, the SH coefficients $\mathbf{c}_i \in \mathbb{R}^{n \times 3}$ and scalar opacity $\alpha_i \in \mathbb{R}$ are the primary components that determine the high-frequency visual details and transparency of the scene, respectively. Our framework focuses on the secure manipulation of these two specific attributes to achieve robust information embedding and recovery.

\section{Methodology}
\label{sec:method}

\subsection{Overview}
The overall workflow of our framework is illustrated in Fig.~\ref{fig2}. Our goal is to embed hidden 3D content into a 3DGS asset while fully preserving the vanilla attribute format and rendering pipeline, so that standard usage for regular users remains unchanged. We jointly optimize SH coefficients and opacity for original and hidden views while constraining Gaussian locations to be shared. The two components are fused via an opacity gradient-based gate under joint rendering supervision.  With an owner-held private key, the embedded content can be reliably recovered for copyright verification. Concretely, we first optimize a paired set of hidden SH coefficients and opacities for each Gaussian primitive to reconstruct the hidden scene under shared geometry. We then perform importance-graded SH coefficient encryption/decryption together with Hash-Grid Guided Opacity Mapping to fuse the hidden attributes into the public asset and enable deterministic recovery. 


\subsection{An Insight into Spherical Harmonics} \label{insight}
\begin{figure}[t]
  \centering
  \includegraphics[width=1\linewidth]{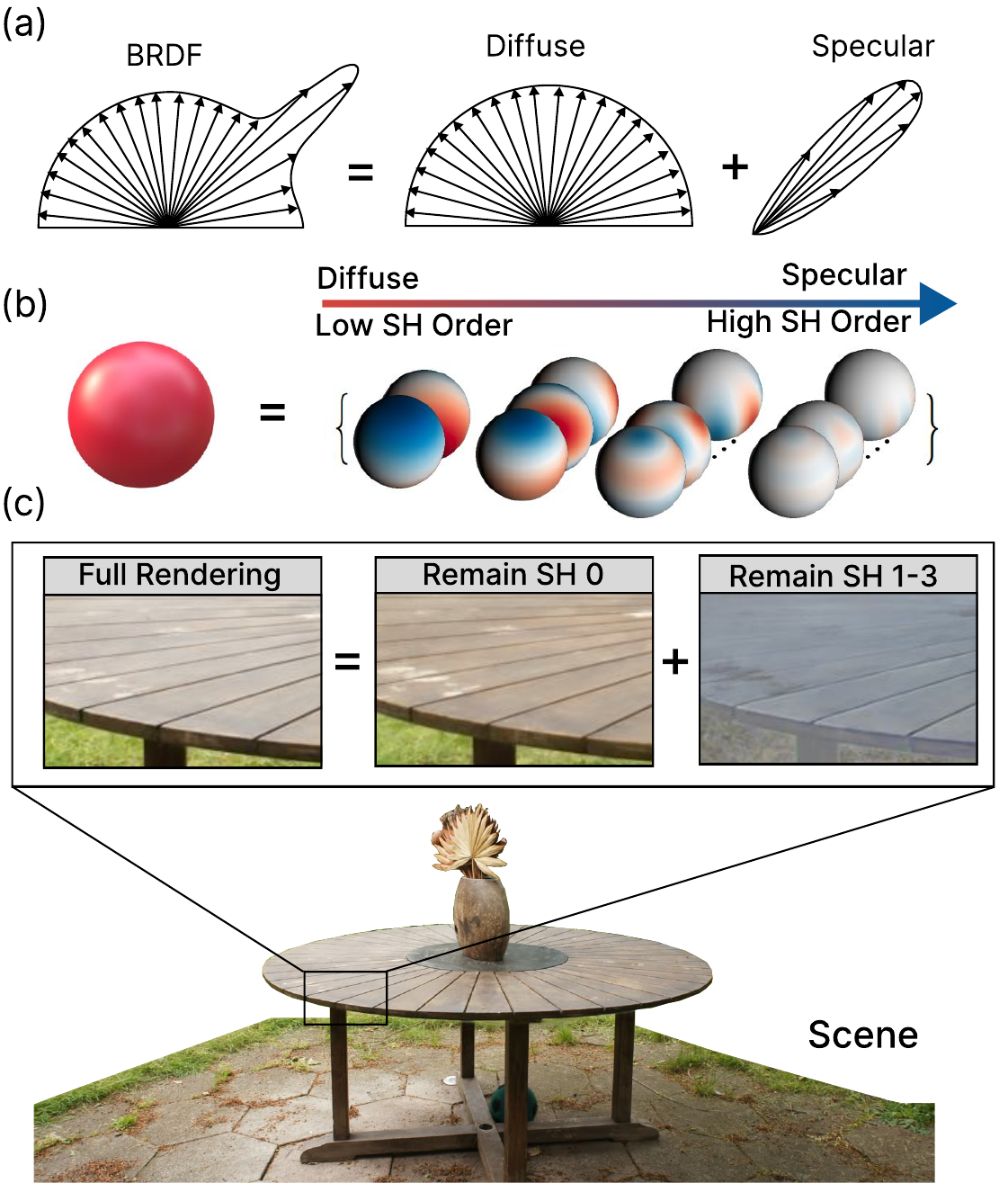}
  \caption{\textbf{(a) Decomposition of scene appearance.} A standard BRDF can be decomposed into diffuse and specular components, with the diffuse component being invariant to the viewing direction, whereas the specular component exhibits pronounced view-dependent behavior. \textbf{(b) Importance of Spherical Harmonics Order.}
We visualize the spherical harmonics (SH) decomposition of the outgoing radiance distribution. Low-order SH terms correspond to view-independent components and dominate the representation, as indicated by darker colors, whereas high-order SH terms encode view-dependent effects and contribute only marginally, as reflected by lighter colors. \textbf{(c) Rendering results with different retained SH orders.} Keeping only SH order 0 or low-order SH preserves most diffuse appearance, while higher-order SH terms mainly contribute to view-dependent specular highlights.}
  \label{fig_SH}
\end{figure}

As established in \cite{sloan2002precomputed}, Spherical Harmonics $Y^m_l$ form an orthonormal basis over the unit sphere $S^2$, analogous to the Fourier transform on a 1D circle. For 3DGS, we utilize a real-valued basis parameterized by $(\theta, \phi)$, where $l \in \mathbb{N}$ denotes the band index and $m \in [-l, l]$ is the frequency order. Low-order bands represent low-frequency components, with the $l$-th band corresponding to polynomials of degree $l$ in Cartesian coordinates $(x, y, z)$.
Any spherical function $F(s)$ (e.g., view-dependent color) can be approximated by a linear combination of SH basis functions:
\begin{equation}
    F(s) \approx \sum_{l=0}^{k-1} \sum_{m=-l}^{l} f_l^m Y_l^m(s) = \sum_{i=0}^{k^2-1} f_i Y_i(s),
    \label{eq:sh_reconstruction}
\end{equation}
where $k$ is the SH degree, $f_i$ are the projection coefficients, and $i = l(l+1)+m$ is the single-index notation. Exploiting the orthonormality of the basis, the coefficients $f_l^m$ are obtained via projection:
\begin{equation}
    f_l^m = \int_{S^2} F(s) Y_l^m(s) ds.
    \label{eq:sh_projection}
\end{equation}
and the basis functions are defined as:
\begin{align}
   Y^m_l(\theta,\phi)=K^m_le^{im\phi}P_l^{|m|},l \in N,m \in (-l,l),
   \label{sh_level}
\end{align}
where $P_l^{|m|}$ are the associated Legendre polynomials and $K^m_l$ are the normalization constants:
\begin{align}
    K^m_l=\sqrt{\frac{(2l+1)(l-|m|)!}{4\pi (l+|m|)!}}.
\end{align}
From Eq.~\ref{sh_level}, it follows that low values of $l$ represent low-frequency 
basis functions over the sphere while high values of $l$ represent high-frequency 
basis functions.
In most cases, higher-frequency reflections occupy only a small proportion of the scene. Consequently, the contribution of higher-order spherical harmonic coefficient is minimal, leading to information redundancy in these higher-order terms, as illustrated in Fig.~\ref{fig_SH}. Embedding information within these terms would pose considerable challenges for detection and would ultimately preserve fidelity. In this paper, we capitalize on this property of spherical harmonics to embed information while simultaneously improving resilience against noise attacks that target spherical harmonics.


\begin{figure*}[t]
  \centering
  \includegraphics[width=0.95\linewidth]{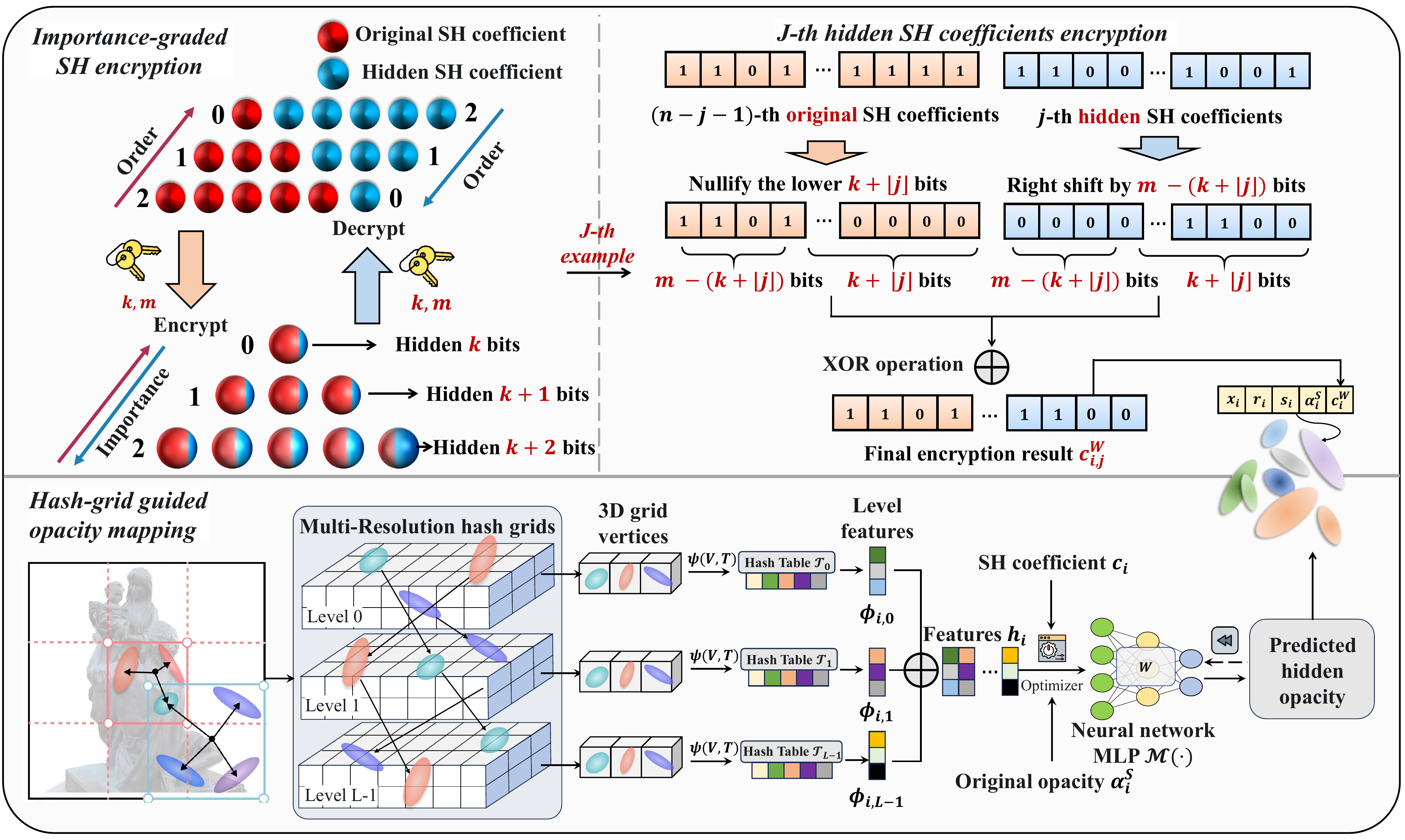}
\caption{\textbf{Overview of the proposed dual-stream steganography framework.} The architecture consists of two key modules: 
\textbf{Top: Importance-graded SH Encryption}, which embeds hidden bit-streams into SH coefficients using an importance-graded ranking strategy. By selectively encrypting SH coefficients across frequency orders via bit-wise XOR operations, it prioritizes the protection of critical visual components. 
\textbf{Bottom: Hash-grid Guided Opacity Mapping}, which regularizes hidden attribute recovery through multi-resolution spatial representations. By retrieving learnable embeddings from hierarchical hash tables and fusing them with primitive attributes, the mapping MLP robustly reconstructs the hidden opacity.}
  \label{fig_shhash}
\end{figure*}

\subsection{3D Information Embedding}
\label{Sec:3D Information Embedding}
Based on the physical insights of 3DGS attributes, we propose a dual-stream steganography framework as illustrated in Fig. \ref{fig_shhash}. The architecture comprises two core components: (i) an \textbf{importance-graded SH encryption} module (Fig. \ref{fig_shhash}, Top) that secures hidden bit-streams within the appearance attributes; and (ii) a \textbf{Hash-Grid Guided Opacity Mapping} module (Fig. \ref{fig_shhash}, Bottom) that robustly reconstructs the hidden geometry. By decoupling the hidden information into graded SH coefficients and a regularized opacity manifold, the framework achieves high-capacity embedding while maintaining exceptional resilience against structural attacks.

\subsubsection{Importance-graded SH Coefficient Encryption}
\label{Sec:Importance-graded SH Coefficient Encryption}
As illustrated in the upper stream of Fig. \ref{fig_shhash}, we embed the hidden SH coefficients $c_i' \in \mathbb{R}^{n \times 3}$ into the original SH coefficients $c_i \in \mathbb{R}^{n \times 3}$. Let $c_{i,j}$ denote the $j$-th component ($0 \leq j \leq n-1$) of $c_i$, whose frequency order is defined as $\lfloor \sqrt{j} \rfloor$. Since lower-order components represent more significant visual features, we draw from the insight in Sec. \ref{insight} to embed more bits of the low-order hidden coefficients $c_{i,j}'$ into the higher-order original coefficients $c_{i,j}$. This strategy maximizes embedding capacity while maintaining visual imperceptibility. To facilitate bit-level manipulation and ensure numerical stability, these coefficients are scaled and represented in a $\gamma$-bit integer format $\{0, 1\}^\gamma$.

The encryption process begins with the \textbf{preparation of the embedding space}. Specifically, we nullify the lower bits of the original SH coefficients based on their graded importance:
\begin{equation}
\tilde{c}_{i,j} = c_{i,j} \enspace \& \sim((1 \ll (k + \lfloor \sqrt{j} \rfloor)) - 1),
\label{eq:SH_null}
\end{equation}
where $\ll$ denotes the left bit-shifting operation, $\sim$ represents the bitwise NOT operation, and $k$ indicates the base bit-shifting length for the zero-order coefficient. 

Building upon this prepared space, we execute the \textbf{bit-wise fusion} by shifting the hidden coefficients to the target bit positions and performing an exclusive OR (XOR) operation. The resulting stego SH coefficients ${c}_{i,j}^{W}$ are formulated as:
\begin{equation}
{c}_{i,j}^{W} = \tilde{c}_{i,j} \oplus ({c}_{i,n-1-j}' \gg (\gamma - (k + \lfloor \sqrt{j} \rfloor))),
\label{eq:SH_fuse}
\end{equation}
where $\gg$ denotes the right bit-shifting operation and $\gamma$ represents the maximum bit length. Notably, the term $n-1-j$ enforces a \textbf{reversed importance order} for embedding. This mechanism ensures that the most critical components of the hidden message are prioritized and anchored within the original high-frequency components, which are inherently less sensitive to human perception.

\subsubsection{Hash-Grid Guided Opacity Mapping}
\label{Sec:Hash-Grid Guided Opacity Mapping}
As illustrated in the lower stream of Fig. \ref{fig_shhash}, the recovery of hidden opacity $\alpha^M$ is grounded in the spatial-attribute coherence of 3D Gaussian primitives. Since the opacity field represents the underlying geometry rather than a collection of independent points, it exhibits significant spatial correlation. Instead of optimizing a naive point-wise mapping, we introduce a multi-resolution hash grid encoder to regularize the transformation. This approach projects the discrete attribute mapping into a continuous latent manifold, ensuring that the reconstructed attributes inherit the geometric priors of the original scene.

Specifically, for a Gaussian primitive $i$ with normalized coordinates $\mathbf{x}_i \in [0, 1]^3$, we establish a hierarchy of $L$ resolution levels to capture multi-scale geometric features. The resolution $R_l$ at each level $l \in \{0, \dots, L-1\}$ is determined by a geometric progression:
\begin{equation}
R_l = \lfloor R_{min} \cdot \eta^l \rfloor, \quad \eta = \exp\left( \frac{\ln R_{max} - \ln R_{min}}{L-1} \right).
\end{equation}
At each level, the coordinate $\mathbf{x}_i$ is scaled by $R_l$ to identify its enclosing voxel cell. Let $\mathcal{V}_l$ denote the eight integer vertices of this cell. We retrieve the corresponding learnable embeddings from a hash table $\mathcal{T}_l$ of size $T$. The mapping from 3D grid vertices to table indices is performed via a fixed spatial hashing function $\Psi$:
\begin{equation}
\Psi(\mathbf{v}, T) = \left( \bigoplus_{d=1}^3 v_d \cdot \pi_d \right) \mod T, \quad \mathbf{v} \in \mathcal{V}_l,
\end{equation}
where $\pi_d$ represents unique large prime numbers. After obtaining the vertex embeddings, we perform trilinear interpolation based on the relative position of $\mathbf{x}_i$ within the cell, yielding a level-specific feature $\phi_{i,l} \in \mathbb{R}^F$. The multi-scale spatial descriptor $\mathbf{h}_i$ is then formed by concatenating these features: 
\begin{equation}
\mathbf{h}_i = [\phi_{i,0}, \dots, \phi_{i,L-1}] \in \mathbb{R}^{L \cdot F}.
\end{equation}

To bridge the gap between spatial geometry and visual appearance, we fuse $\mathbf{h}_i$ with the original scene opacity $\alpha^S_i$ and the primary SH coefficients $\mathbf{c}_i$. The Hash-Grid Guided Opacity Mapping module, parameterized by the hash table embeddings $\mathcal{T}$ and the mapping network weights $\mathbf{W}$, is jointly optimized by minimizing the reconstruction discrepancy:
\begin{equation}
\mathcal{L}_{op}(\mathcal{T}, \mathbf{W}) = \sum_{i=1}^{N} \left\| \mathcal{M} \left( \mathbf{h}_i \oplus \alpha^S_i \oplus \mathbf{c}_i ; \mathbf{W} \right) - \alpha^M_i \right\|^2,
\end{equation}
where $\oplus$ denotes the concatenation operator. By encapsulating the spatial-attribute mapping into $\{\mathcal{T}, \mathbf{W}\}$, the framework enables the MLP $\mathcal{M}$ to infer hidden opacity values through neural interpolation. This mechanism exploits multi-scale spatial priors to facilitate high-fidelity recovery of the hidden structure while maintaining exceptional parameter efficiency.

\subsection{3D Information Extraction}
\label{Sec:3D Information Extraction}
The extraction process is a deterministic recovery of hidden attributes from the stego 3DGS asset. Given the secret key consisting of the bit-shifting parameters and the optimized mapping parameters $\{\mathcal{T}^*, \mathbf{W}^*\}$, the estimated hidden SH coefficients $\hat{\mathbf{c}}_i^M$ and opacity $\hat{\alpha}_i^M$ are reconstructed through bit-domain decryption and neural manifold inference.

For each Gaussian primitive $i$, the hidden SH coefficients are retrieved by isolating the encapsulated bit-planes from the fused coefficients $\mathbf{c}_{i,j}^{W}$:
\begin{equation}
\hat{c}_{i,j}^M = c_{i,n-1-j}^{W} \enspace \& \enspace (1 \ll (k + \lfloor \sqrt{n-1-j} \rfloor)).
\label{eq:sh_recovery}
\end{equation}
This operation extracts the message bits anchored within the original high-frequency components based on the reversed importance order.

Simultaneously, the hidden opacity is reconstructed by feeding the spatial-attribute features into the pre-trained Hash-Grid Guided Opacity Mapping module. The recovery of $\hat{\alpha}^M_i$ is computed via a forward pass of the mapping network $\mathcal{M}$ over the primitive set:
\begin{equation}
\hat{\alpha}^M_{i} = \mathcal{M} \left( \Phi(\mathbf{x}_i; \mathcal{T}^*) \oplus \alpha^S_i \oplus \mathbf{c}^S_i ; \mathbf{W}^* \right),
\label{eq:op_recovery}
\end{equation}
where $\Phi(\cdot; \mathcal{T}^*)$ denotes the multi-resolution spatial encoding. This inference mechanism ensures that the reconstructed opacity $\hat{\alpha}^M_i$ inherits the geometric consistency of the hash grid and the appearance priors of the original Gaussian representation.

The recovered attributes $\{\mathbf{x}_i, \mathbf{r}_i, \mathbf{s}_i, \hat{\alpha}_i^M, \hat{\mathbf{c}}_i^M\}$ are integrated to reconstruct the hidden 3DGS model. The decoded content is rendered via standard rasterization, facilitating high-fidelity visualization and robust copyright authentication.

\subsection{Training Details}  
To train our method, we assume access to paired training data consisting of original scenes
$\{\mathbf{I}_{\mathrm{gt}}^{S}\}$ and corresponding hidden scenes (message)
$\{\mathbf{I}_{\mathrm{gt}}^{M}\}$.
The learnable parameters of our model are denoted by
$\boldsymbol{\Theta}_i = \{x_i, r_i, s_i, \alpha_i^S, \mathbf{c}_i^S,\alpha_i^M, \mathbf{c}_i^M\}$.
Here, $\mathbf{x}_i$, $\mathbf{r}_i$, and $\mathbf{s}_i$ denote the shared parameters across both representations,
while $\alpha_i^{S}$ and $\mathbf{c}_i^{S}$ are used to model the original scene,
and $\alpha_i^{M}$ and $\mathbf{c}_i^{M}$ are employed to represent the hidden message, respectively.
Following the standard formulation of 3DGS, the training objective of our method is defined by a combination of image reconstruction
losses for both the original scene and the hidden message scene.
Specifically, the original scene reconstruction loss is formulated as
\begin{equation}
\label{eq:rgb_loss}
\ell_{\mathrm{scnen}}
=
(1 - \gamma)\,\ell_1(\mathbf{I}^S_{\mathrm{pred}}, \mathbf{I}^S_{\mathrm{gt}})
+ \gamma\,\ell_{\mathrm{SSIM}}(\mathbf{I}^S_{\mathrm{pred}}, \mathbf{I}^S_{\mathrm{gt}}),
\end{equation}

Similarly, the reconstruction loss for the hidden message scene is defined as
\begin{equation}
\label{eq:message_loss}
\ell_{\mathrm{mes}}
=
(1 - \beta)\,\ell_1(\mathbf{I}^M_{\mathrm{pred}}, \mathbf{I}^M_{\mathrm{gt}})
+ \beta\,\ell_{\mathrm{SSIM}}(\mathbf{I}^M_{\mathrm{pred}}, \mathbf{I}^M_{\mathrm{gt}}),
\end{equation}
where $\mathbf{I}^S_{\mathrm{pred}}$ and $\mathbf{I}^M_{\mathrm{pred}}$ denote the rendered images for original scene and hidden scene respectively.
The overall training objective is then given by
\begin{equation}
\label{eq:total_loss}
\ell_{\mathrm{recon}} = \ell_{\mathrm{scene}} + \mathrm{\lambda^M}\ell_{\mathrm{mes}},
\end{equation}
where $\lambda^M$ balances the optimization between preserving the visual fidelity
of the original scene and accurately reconstructing the hidden message scene.
\begin{figure*}[t]
  \centering
  \includegraphics[width=1.0\linewidth]{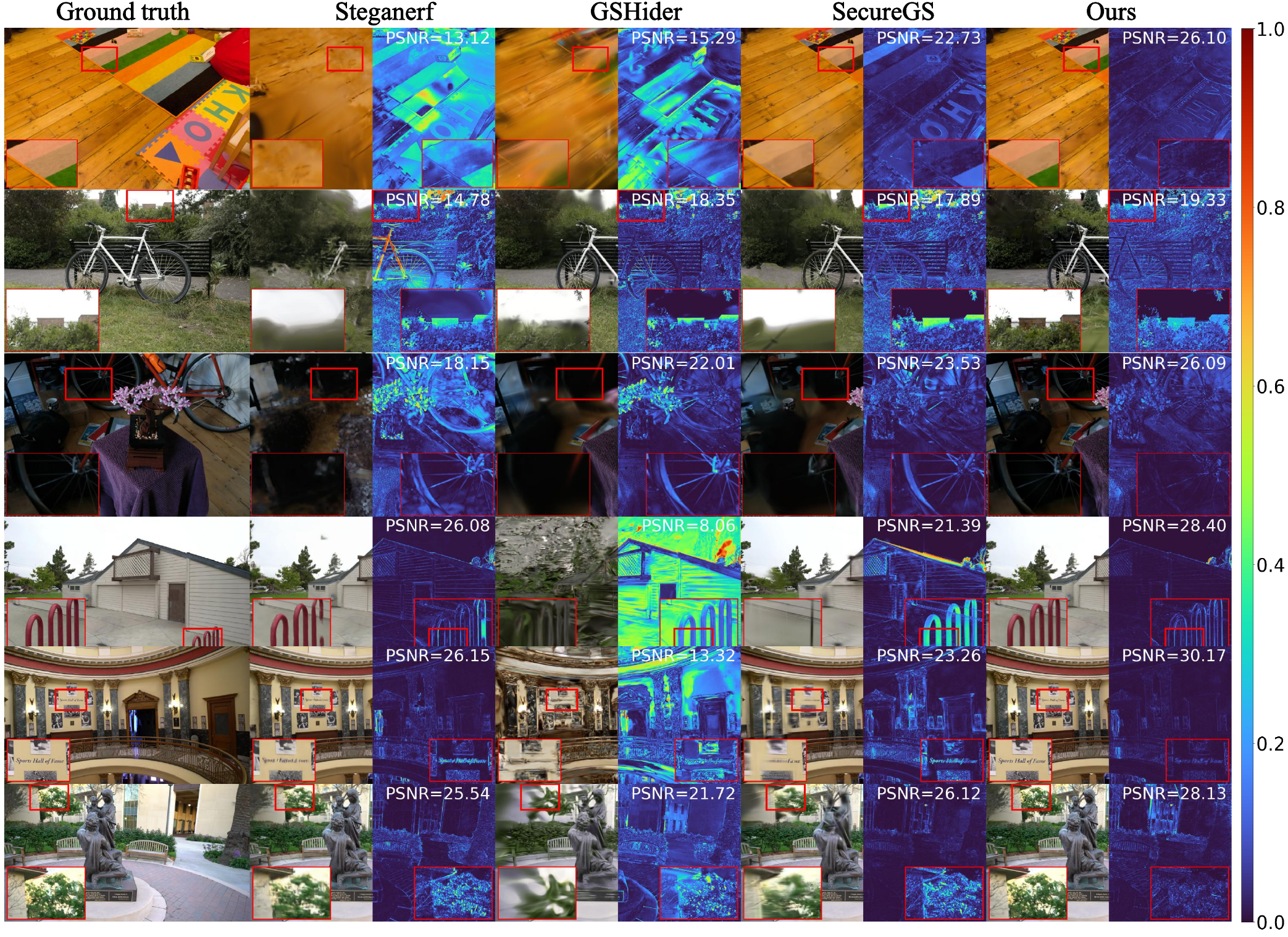}
  \caption{Qualitative comparisons. The difference maps visualize the pixel-wise residuals between GT and renderings with 10× amplification for better
visibility Our results achieved sharper, more consistent structures without the artifacts seen in GS-Hider or the erroneous textures and colors observed in StegaNeRF+GS.}
  \label{fig_compare}
\end{figure*}
During optimization, we associate each Gaussian primitive with two opacities, denoted as $\alpha^{S}$ and $\alpha^{M}_i$, corresponding to the scene and message, respectively. Employing two independent opacity variables to represent the scene and the embedded message separately helps mitigate geometric ambiguity and potential information leakage. However, such a naive optimization strategy may drive excessive divergence between the opacity fields, resulting in geometric separation of the original and embedded scenes and increased susceptibility to adversarial manipulation.
To address this issue and facilitate structural coupling, we propose a visibility- and gradient-aware opacity consistency regularization that selectively enforces agreement only on stable and informative Gaussians.
This design encourages the scene and message to share a coherent geometric support, thereby strengthening their coupling and improving robustness against adversarial perturbations.
Specifically, we model opacity as the parameter of a Bernoulli distribution and measure the discrepancy between $\alpha^{S}_i$ and $\alpha^{M}_i$ using a symmetric Kullback--Leibler (KL) divergence.
The consistency loss is modulated by two factors: a visibility weight $v_i$ derived from volumetric rendering, and a gradient-based gate
\begin{equation}
g_i = \exp\!\left(- \left\| \nabla_{{\alpha}_i^S} \mathcal{L}_{\mathrm{recon}} \right\| \right),
\end{equation}
This gating mechanism relaxes the constraint in regions undergoing active appearance optimization, while enforcing stronger alignment in well-observed and stable regions.

The final opacity consistency loss is defined as
\begin{equation}
\mathcal{\ell}_{\mathrm{cons}}
=
\mathbb{E}_i
\left[
g_i \, v_i \,
D_{\mathrm{KL}}^{\mathrm{sym}}\!\left(
\mathrm{Bern}(\alpha_i^{S}) \,\|\, \mathrm{Bern}(\alpha_i^{M})
\right)
\right].
\end{equation}
By coupling the scene and message through a shared and adaptive opacity structure, this formulation promotes a tightly fused representation that is less susceptible to structural disruption, thereby enhancing resistance to attacks.
The overall training objective is then given by
\begin{equation}
\label{eq:total_loss2}
\ell_{\mathrm{total}} = \ell_{\mathrm{recon}} + \lambda\ell_{\mathrm{cons}},
\end{equation}
where $\lambda$ balances the optimization between preserving the visual fidelity and preserving the robustness.

\begin{table*}[t]
\setlength{\tabcolsep}{2mm}
\centering
\resizebox{1.0\linewidth}{!}{
\begin{tabular}{c||c|cc|cc|cc|cc}
\hline\thickhline
\rowcolor[HTML]{e6f0f0}
 & 
& \multicolumn{2}{c|}{\textbf{3DGS+StegaNeRF}}
& \multicolumn{2}{c|}{\textbf{GS-Hider}}
& \multicolumn{2}{c|}{\textbf{SecureGS}}
& \multicolumn{2}{c}{\textbf{Ours}} \\
\rowcolor[HTML]{e6f0f0} \multirow{-2}{*}{\textbf{Dataset}}
& \multirow{-2}{*}{\textbf{Metric}}
& \textbf{Scene} & \textbf{Message}
& \textbf{Scene} & \textbf{Message}
& \textbf{Scene} & \textbf{Message}
& \textbf{Scene} & \textbf{Message} \\
\hline\hline

\multirow{3}{*}{Mip-Nerf360}
& PSNR $\uparrow$
& 24.120 & 16.681 & 25.817 & \second{25.179} & \second{26.574} & 23.679 & \best{26.694} & \best{27.555} \\
& SSIM $\uparrow$
& 0.763 & 0.538 & \second{0.782} & \second{0.780} & 0.775 & 0.734 & \best{0.782} & \best{0.829} \\
& LPIPS $\downarrow$
& 0.319 & 0.574 & \best{0.246} & \second{0.306} & \second{0.255} & 0.388 & 0.257 & \best{0.265} \\
\hline

\multirow{3}{*}{Tanks \& Temples}
& PSNR $\uparrow$
& 27.251 & \best{28.114} & 26.010 & 24.388 & \second{27.865} & 22.632 & \best{29.027} & \second{27.381} \\
& SSIM $\uparrow$
& \second{0.888} & \second{0.860} & 0.850 & 0.806 & 0.876 & 0.696 & \best{ 0.904} & \best{ 0.919} \\
& LPIPS $\downarrow$
& \second{0.139} & \second{0.181} & 0.201 & 0.247 & 0.150 & 0.370 & \best{0.125} & \best{ 0.109} \\
\hline

\multirow{3}{*}{NeRF-Synthetic}
& PSNR $\uparrow$
& 30.54 & 20.18 & 31.99 & 21.29 & \second{33.11} & \second{24.96} & \best{33.48} & \best{32.38} \\
& SSIM $\uparrow$
& 0.960 & 0.852 & 0.972 & 0.867 & \second{0.974} & \second{0.905} & \best{0.976} & \best{0.970} \\
& LPIPS $\downarrow$
& 0.049 & 0.161 & 0.031 & 0.149 & \second{0.027} & \second{0.115} & \best{0.025} & \best{0.035} \\
\hline

\multirow{3}{*}{Average}
& PSNR $\uparrow$
& 27.304 & 21.658 & 27.939 & 23.619 & \second{29.183} & \second{23.757} & \best{29.693} & \best{30.037} \\
& SSIM $\uparrow$
& 0.870 & 0.750 & 0.868 & \second{0.818} & \second{0.875} & 0.778 & \best{0.887} & \best{0.900} \\
& LPIPS $\downarrow$
& 0.169 & 0.305 & 0.159 & \second{0.234} & \second{0.144} & 0.291 & \best{0.140} & \best{0.147} \\
\hline

\end{tabular}
}
\caption{Quantitative comparison on four datasets using average \textbf{PSNR/SSIM/LPIPS} for both the \textbf{scene} (public content) and the \textbf{message} (hidden content). \best{Red bold} and \second{blue} indicate the best and second-best performances, respectively.}
\label{tab_main}
\end{table*}

\section{Experiment}

\subsection{Implementation Details}

Our framework is implemented in PyTorch and optimized on an NVIDIA A800 GPU. The 3DGS baseline is trained for 30,000 iterations with balancing weights $\lambda^M=1.0$ and $\lambda=0.02$, following the standard optimization schedule, while all other hyperparameters remain consistent with the vanilla 3DGS implementation. For the Hash-Grid Guided Opacity Mapping module, we employ a configuration with $L=16$, $R_{min}=16$, $R_{max}=1024$, $T=2^{21}$, and $F=4$. The mapping network and hash tables are jointly optimized for 4,000 epochs using the Adam optimizer with a learning rate of $5 \times 10^{-3}$. The bit-shifting parameter $k$ for SH coefficient encryption is set to 17.

\subsection{Experimental Settings}
\subsubsection{Datasets} For 3D Steganography, we evaluate our method on three established datasets commonly used for novel view synthesis, including the NeRF synthetic dataset~\cite{nerf}, the Tanks and Temples dataset~\cite{knapitsch2017tanks}, and the Mip-NeRF360 dataset~\cite{barron2021mipnerf}. The  NeRF synthetic dataset consists of eight detailed synthetic objects, each rendered from 100 virtual camera views uniformly distributed on an inward-facing hemisphere. The Tanks and Temples dataset includes realistic scenes with large central objects, where each scene contains between 263 and 1,107 images captured using a monocular RGB camera. The Mip-NeRF360 dataset consists of nine indoor and outdoor scenes featuring complex central objects. Following the experimental settings of ~\cite{zhang2024gshider,zhang2025securegs}, for each dataset, we use a different scene from the same dataset as the hidden scene for every original scene.
 For 4D steganography, we evaluate our method on synthetic and real-world datasets, including the D-NeRF dataset~\cite{pumarola2021dnerf} and the HyperNeRF dataset~\cite{park2021hypernerf}. The D-NeRF dataset is designed for monocular capture settings and contains dynamic scenes with approximately randomly sampled camera poses at each timestamp, where each scene consists of between 50 and 200 frames. The HyperNeRF dataset is captured using one or two cameras following relatively simple camera trajectories.

\subsubsection{Baselines} We compare our method with existing 3DGS steganography method GS-Hider \cite{zhang2024gshider} and SecureGS \cite{zhang2025securegs}. 
Meanwhile, we feed the rendering results of the original 3DGS to a U-shaped decoder and constrain it to output hidden scenes according to \cite{li2023steganerf}, called 3DGS+StegaNeRF. For robustness evaluation, we first adapt conventional message purification techniques for comparative analysis, namely opacity pruning, where primitives with lower opacity are progressively eliminated. In addition, we incorporate the latest pruning method specifically designed for 3DGS Steganography, GSPure \cite{huang2025can}, for further robustness evaluation of our approach. 

\begin{figure*}[t]
  \centering
  \includegraphics[width=0.98\linewidth]{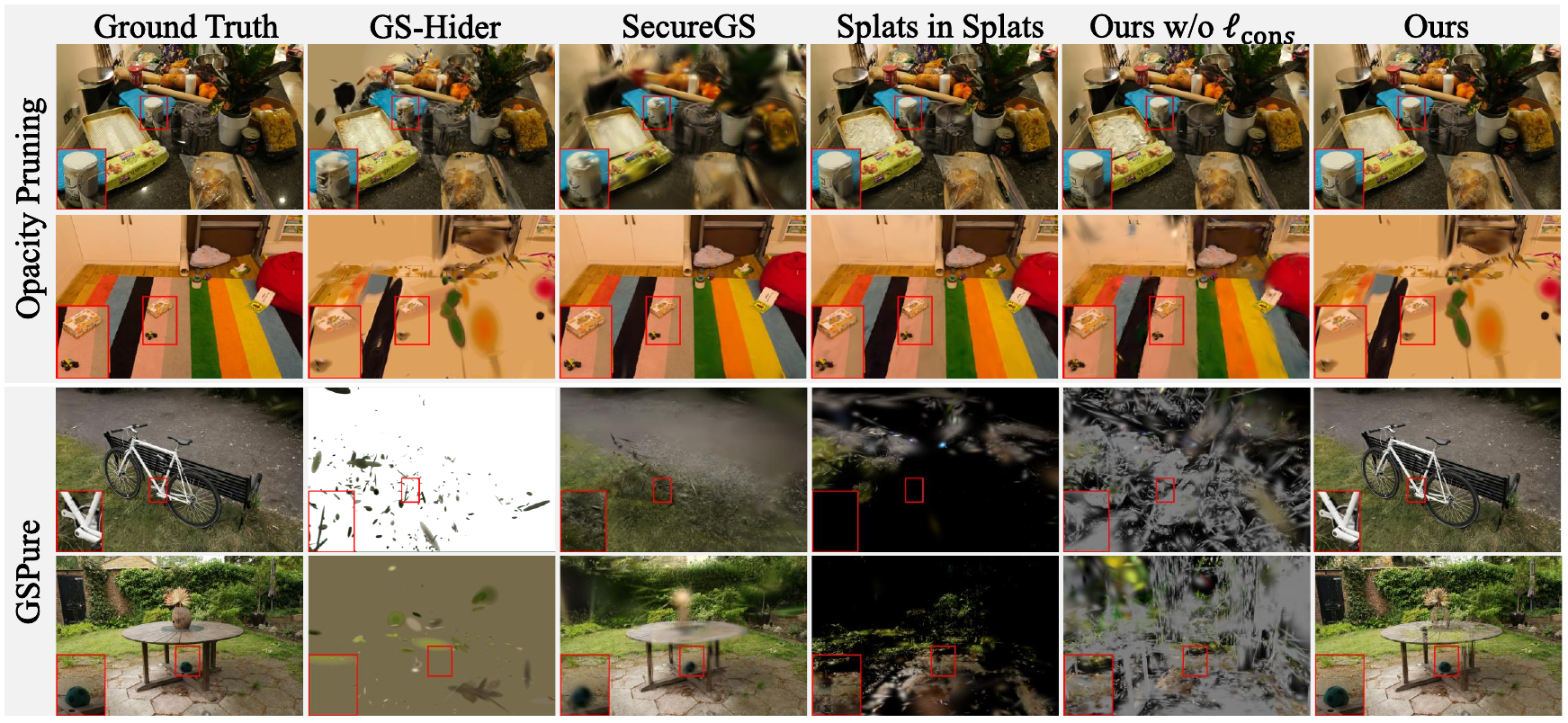}
\caption{\textbf{Visual results under various pruning attacks.} 
Compared with SOTA steganographic frameworks, our method exhibits remarkable robustness against both standard \textit{Opacity Pruning} and the more aggressive \textit{GSPure} attack. Our framework preserves sharp textures and accurate color representation, whereas competing methods show significant fidelity loss or erroneous rendering.}
\label{fig:vis_robustness}
\end{figure*}

\subsubsection{Evaluation Metrics} 
To quantitatively assess the visual fidelity of the synthesized novel views, we employ three standard metrics: Peak Signal-to-Noise Ratio (PSNR), Structural Similarity Index (SSIM), and Learned Perceptual Image Patch Similarity (LPIPS). Additionally, rendering efficiency is measured by Frames Per Second (FPS).
To specifically evaluate the resilience of our framework against pruning attacks, we introduce a \textbf{Robustness Score} ($\mathcal{S}_{R}$), which characterizes the relative persistence of the hidden information compared to the main scene. We define the relative performance decay rate $\mathcal{D}_{rel}$ for a specific metric $M$ as:
\begin{equation}
    \mathcal{D}_{rel}(M) = \frac{|M_{orig} - M_{pruned}|}{M_{orig}}.
\label{eq:relative_decay_simple}
\end{equation}
The Robustness Score is then formulated as the differential between the decay rate of the main scene and that of the hidden message:
\begin{equation}
    \mathcal{S}_{R} = \mathcal{D}_{rel}(M)_{scene} - \mathcal{D}_{rel}(M)_{message}.
\label{eq:robustness_score_simple}
\end{equation}
Under this definition, $\mathcal{S}_{R}$ reflects the "damage ratio" gap between the public and hidden domains. A \textbf{higher} score (closer to zero or positive) indicates that the hidden watermark is more resilient than the scene itself.  

\subsection{Evaluation on 3D Steganography}   
Firstly, we compare Splats in Splats++ with baselines in terms of fidelity, efficiency, robustness, security, and usability.

\subsubsection{Fidelity} We use Novel View Synthesis (NVS) results to evaluate scene fidelity. As shown in Tab.~\ref{tab_main}, our method achieves the best NVS quality on both original and hidden scenes. Fig.~\ref{fig_compare} shows more visual details, which indicate that our method achieves the most appealing visualization results without the artifacts seen in SecureGS or the erroneous textures and colors observed in GS-Hider. The difference maps shown in the right half of Fig.~\ref{fig_compare} more clearly demonstrate that our results consistently exhibit smaller deviations from the ground truth across all scenes.

\subsubsection{Efficiency} Tab.~\ref{tab_eff} demonstrates that our Splats in Splats++ achieves state-of-the-art rendering and
 training efficiency. Our method achieves 100 rendering FPS, which is 3x faster than the existing 3DGS steganography methods using neural networks, such as GS-Hider. Our training time is only half of the previous methods.
 
 \begin{table}[h]
\setlength{\tabcolsep}{1.1mm}
  \centering
    \resizebox{1.0\linewidth}{!}
    {

     \begin{tabular}{ccccp{0.05cm}cc}
\hline\thickhline
\rowcolor[HTML]{e6f0f0}    & \multicolumn{3}{c}{Usability}  & &\multicolumn{2}{c}{Efficiency}\\
 \cline{2-4}  \cline{6-7}
\rowcolor[HTML]{e6f0f0} \multirow{-2}{*}{Method}& Pipe.& Attr. & Viewer & & Train time& FPS\\
\hline

  \centering GS+StegaNeRF  &\XSolidBrush&\Checkmark&\XSolidBrush&&  97 Mins &  22  \\ 
  \centering  SecureGS &\XSolidBrush&\XSolidBrush&\XSolidBrush&&  71 Mins&  36  \\
  \centering GS-Hider   &\XSolidBrush&\XSolidBrush&\XSolidBrush&&  119 Mins&  44 \\
     \centering Ours &\Checkmark&\Checkmark&\Checkmark&&  47 Mins &  118 \\
     \hline
    \end{tabular}%
}
  \caption{Comparison of the usability and efficiency. Pipe. and Atrr. represent maintaining vanilla 3DGS's rendering pipeline and attributes. 
 }
  \label{tab_eff}
\end{table}%

\subsubsection{Usability} Tab.~\ref{tab_eff} also demonstrates the adaptability of Splats in Splats++ and baselines to the rendering engine (SIBR Viewer) provided by 3DGS. Our method can be directly integrated into the rendering pipeline provided by 3DGS, offering a seamless user experience that existing approaches are unable to achieve. We also demonstrate whether all methods maintain the vanilla 3DGS's rendering pipeline (Pipe. in short) and attributes (Attr. in short). 

\begin{table*}[t]
\centering
\small
\setlength{\tabcolsep}{4.0pt}
\renewcommand\arraystretch{1.1}
\resizebox{0.98\linewidth}{!}{
\begin{tabular}{c||ccc|ccc|ccc}
\hline\thickhline
\rowcolor[HTML]{e6f0f0}
& \multicolumn{3}{c|}{\textbf{PSNR} $\uparrow$}
& \multicolumn{3}{c|}{\textbf{SSIM} $\uparrow$}
& \multicolumn{3}{c}{\textbf{LPIPS} $\downarrow$} \\
\cline{2-10}
\rowcolor[HTML]{e6f0f0}
\multirow{-2}{*}{\textbf{Attack}}
& \textbf{Scene} & \textbf{Message} & \textbf{$\mathcal{S_R}$}
& \textbf{Scene} & \textbf{Message} & \textbf{$\mathcal{S_R}$}
& \textbf{Scene} & \textbf{Message} & \textbf{$\mathcal{S_R}$} \\
\hline\hline

{\textcolor{gray!80}{\textit{\textbf{GS-Hider}}}}
& 25.817 & 25.179 & 0.00
& 0.782 & 0.780 & 0.00
& 0.246 & 0.306 & 0.00 \\
\hdashline
Opacity Pruning
& 25.759 & 16.043 & -36.06
& 0.776 & 0.599 & -22.44
& 0.270 & 0.519 & -59.85 \\

GSPure
& 24.808 & 16.139 & -31.99
& 0.767 & 0.617 & -18.98
& 0.278 & 0.494 & -48.43 \\
\hline

{\textcolor{gray!80}{\textit{\textbf{SecureGS}}}}
& 26.574 & 23.679 & 0.00
& 0.775 & 0.734 & 0.00
& 0.255 & 0.388 & 0.00 \\
\hdashline
Opacity Pruning
& 26.574 & 22.380 & \second{-5.49}
& 0.775 & 0.716 & \second{-2.45}
& 0.255 & 0.403 & \second{-3.87} \\

GSPure
& 25.268 & 16.712 & \second{-24.51}
& 0.759 & 0.566 & \second{-20.82}
& 0.275 & 0.525 & \second{-27.47} \\
\hline

{\textcolor{gray!80}{\textit{\textbf{Splats in Splats}}}}
& 26.749 & 26.517 & 0.00
& 0.784 & 0.797 & 0.00
& 0.262 & 0.308 & 0.00 \\
\hdashline
Opacity Pruning
& 26.749 & 24.599 & -7.23
& 0.784 & 0.769 & -3.51
& 0.262 & 0.341 & -10.71 \\

GSPure
& 25.621 & 10.193 & -57.34
& 0.771 & 0.289 & -62.08
& 0.274 & 0.655 & -108.08 \\
\hline


{\textcolor{gray!80}{\textit{\textbf{Ours w/o $\ell_{\mathrm{cons}}$}}}}
& 26.749 & 27.025 & 0.00
& 0.784 & 0.814 & 0.00
& 0.262 & 0.294 & 0.00 \\
\hdashline
Opacity Pruning
& 26.749 & 20.449 & -24.33
& 0.784 & 0.668 & -17.94
& 0.262 & 0.418 & -42.18 \\

GSPure
& 25.619 & 9.537 & -60.49
& 0.771 & 0.239 & -68.98
& 0.275 & 0.727 & -142.32 \\
\hline
{\textcolor{gray!80}{\textit{\textbf{Ours}}}}
& 26.694 & 27.556 & 0.00
& 0.782 & 0.829 & 0.00
& 0.257 & 0.265 & 0.00 \\
\hdashline
Opacity Pruning
& 26.694 & 27.252 & \best{-1.10}
& 0.782 & 0.824 & \best{-0.60}
& 0.257 & 0.268 & \best{-1.13} \\

GSPure
& 25.431 & 22.185 & \best{-14.76}
& 0.770 & 0.741 & \best{-9.08}
& 0.271 & 0.337 & \best{-21.72} \\
\hline\thickhline
\end{tabular}
}
\caption{
\textbf{Robustness under pruning-based watermark removal attacks.}
    The Robustness Score $\mathcal{S}_{R}$ evaluates the gap between scene and message decay rates. In practice, $\mathcal{S}_{R}$ is typically negative as the hidden message is inherently more susceptible to pruning-induced degradation than the broader scene geometry. Our method aims to maximize $\mathcal{S}_{R}$ (driving it toward 0), which signifies that the hidden manifold remains as stable as the main scene during structural attacks, effectively preventing disproportionate watermark loss. \best{Red bold} and \second{blue} indicate the best and second-best performances, respectively. Our method demonstrates superior resilience across all pruning strategies.}
\label{tab:pruning_attacks_summary}
\end{table*}

\subsubsection{Robustness Analysis} 
To evaluate the resilience of our framework in practical scenarios, we conduct extensive stress tests against structural pruning, the most prevalent degradation for 3DGS assets. We benchmark our method under two distinct regimes: \textit{Opacity Pruning}, which eliminates primitives based on importance rankings, and \textit{GSPure}, an advanced compression strategy targeting significant Gaussians. As summarized in Tab.~\ref{tab:pruning_attacks_summary}, while competing methods exhibit catastrophic fidelity loss under structural attacks, our framework maintains exceptional recovery stability. Specifically, under Opacity Pruning, our method achieves a Message PSNR of $\bm{27.252}$ dB, yielding a Robustness Score of $\bm{-1.10}$ that significantly outperforms the second-best framework ($\bm{-5.49}$). In the more aggressive GSPure attack, our approach sustains a high-fidelity recovery with $\bm{22.185}$ dB in Message PSNR, whereas the second-best method suffers a severe performance drop to $\bm{-24.51}$ in Robustness Score. 
This superior robustness is visually confirmed in Fig.~\ref{fig:vis_robustness}, where our rendered scenes remain structurally intact and sharp even under extreme pruning, while existing methods show fragmented artifacts or geometric collapse. Our framework provides a resilient solution for copyright persistence, effectively bridging the gap between asset optimization and structural robustness.
 Such resilience stems from the density-gating mechanism, which effectively couples the original scene with the message. Detailed empirical validation is provided in the ablation study.

\subsubsection{Security} 
\begin{figure}[h]
  \centering
    \includegraphics[width=\linewidth]{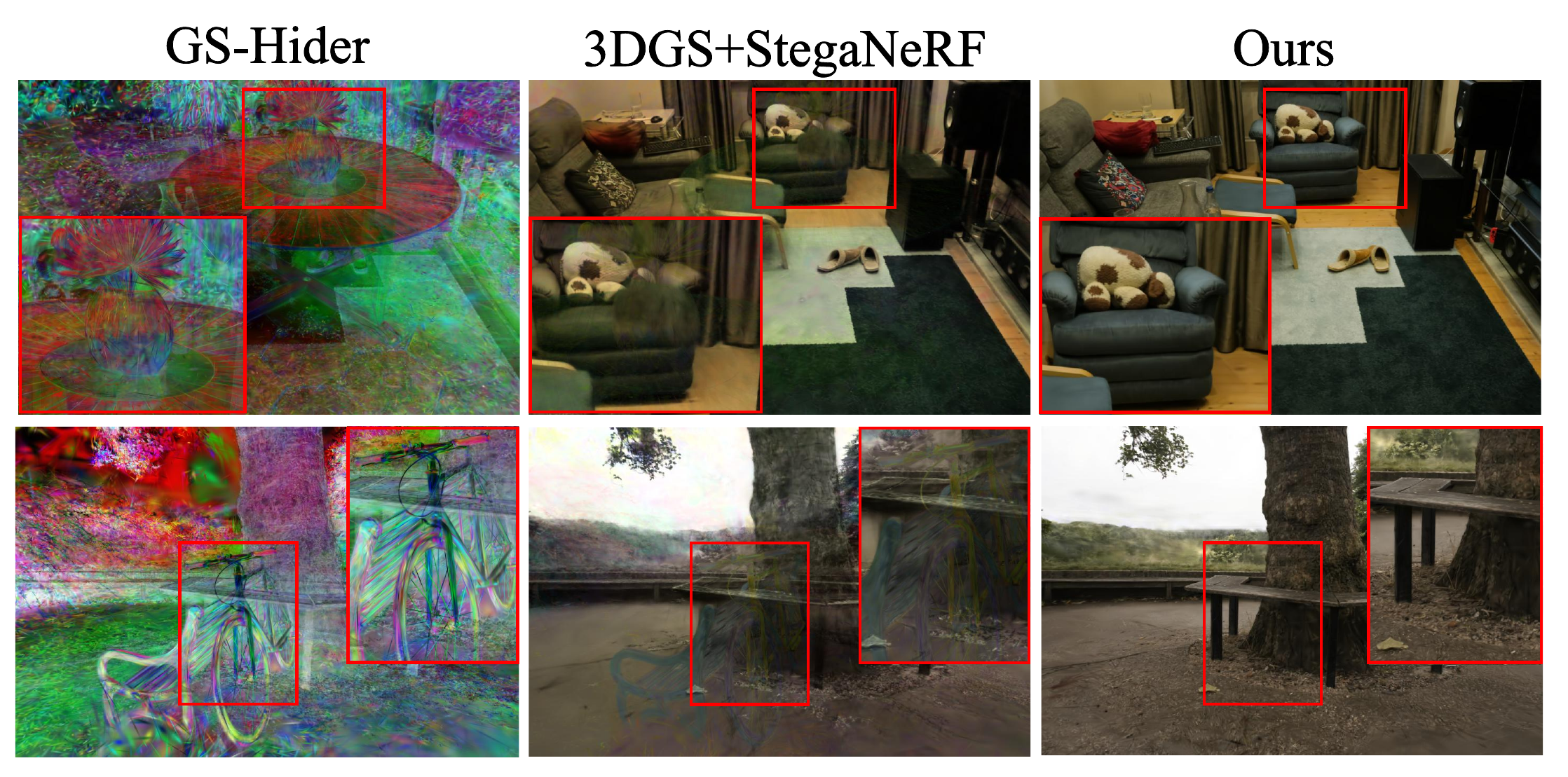}
    \caption{Comparison of rendering results. GS-hider and 3DGS+SteganeNeRF have obvious information leakage. \textbf{Better viewed on screen with zoom in}.}
    \label{fig_roc}
\end{figure}

We analyze the security from the following aspects. From the view of point cloud file, it preserves the same structure as the original 3DGS. From the view of output, our scenes do not exhibit any artifacts or edges of hidden scenes and maintain high fidelity (shown in Fig.~\ref{fig_compare} and Tab.~\ref{tab_main}). Finally, we visualize the rendering results which can also be directly obtained by users.  Fig.~\ref{fig_roc} indicates that the rendering results of GS-Hider and 3DGS+StegaNeRF have obvious hidden information leakage while our method ensure the security.

\begin{table*}[t]
\setlength{\tabcolsep}{2mm}
\centering
\resizebox{1.0\linewidth}{!}{
\begin{tabular}{c||c|cc|cc|cc|cc}
\hline\thickhline
\rowcolor[HTML]{e6f0f0}
 & 
& \multicolumn{2}{c|}{\textbf{3DGS+StegaNeRF}}
& \multicolumn{2}{c|}{\textbf{GS-Hider}}
& \multicolumn{2}{c|}{\textbf{SecureGS}}
& \multicolumn{2}{c}{\textbf{Ours}} \\
\rowcolor[HTML]{e6f0f0} \multirow{-2}{*}{\textbf{Dataset}}
& \multirow{-2}{*}{\textbf{Metric}}
& \textbf{Scene} & \textbf{Message}
& \textbf{Scene} & \textbf{Message}
& \textbf{Scene} & \textbf{Message}
& \textbf{Scene} & \textbf{Message} \\
\hline\hline

\multirow{3}{*}{DNeRF}
& PSNR $\uparrow$
& 23.589 & 14.709 & 23.047 & 25.687 & \second{29.740} & \second{28.068} & \best{33.318} & \best{29.368} \\
& SSIM $\uparrow$
& 0.926 & 0.696 & 0.919 & 0.905 & \second{0.950} & \second{0.928} & \best{0.972} & \best{0.940} \\
& LPIPS $\downarrow$
& 0.106 & 0.311 & 0.101 & 0.106 & \second{0.064} & \second{0.071} & \best{0.024} & \best{0.058} \\
\hline

\multirow{3}{*}{HyperNeRF}
& PSNR $\uparrow$
& 18.698 & 12.452 & 19.651 & 16.725 & \second{ 22.910} & \second{19.880} & \best{24.893} & \best{ 24.209} \\
& SSIM $\uparrow$
& 0.609 & 0.415 & \second{0.670} & 0.441 & 0.630 & \second{ 0.552} & \best{0.678} & \best{ 0.731} \\
& LPIPS $\downarrow$
& 0.512 & 0.521 & 0.420 & 0.633 & \second{0.400} & \second{0.505} & \best{0.303} & \best{0.366} \\
\hline


\multirow{3}{*}{Average}
& PSNR $\uparrow$
& 21.143 & 13.580 & 21.349 & 21.206 & \second{26.325} & \second{23.974} & \best{31.397} & \best{26.789} \\
& SSIM $\uparrow$
& 0.767 & 0.555 & 0.510 & 0.673 & \second{0.790} & \second{0.740} & \best{0.825} & \best{0.864} \\
& LPIPS $\downarrow$
& 0.309 & 0.416 & 0.260 & 0.369 & \second{0.232} & \second{0.288} & \best{0.164} & \best{0.188} \\
\hline

\end{tabular}
}
\caption{Quantitative comparison on 4D datasets using average \textbf{PSNR/SSIM/LPIPS} for both the \textbf{scene} (public content) and the \textbf{message} (hidden content). \best{Red bold} and \second{blue} indicate the best and second-best performances, respectively.}
\label{tab:4d_results}
\end{table*}
\begin{figure*}[t]
  \centering
  \includegraphics[width=0.98\linewidth]{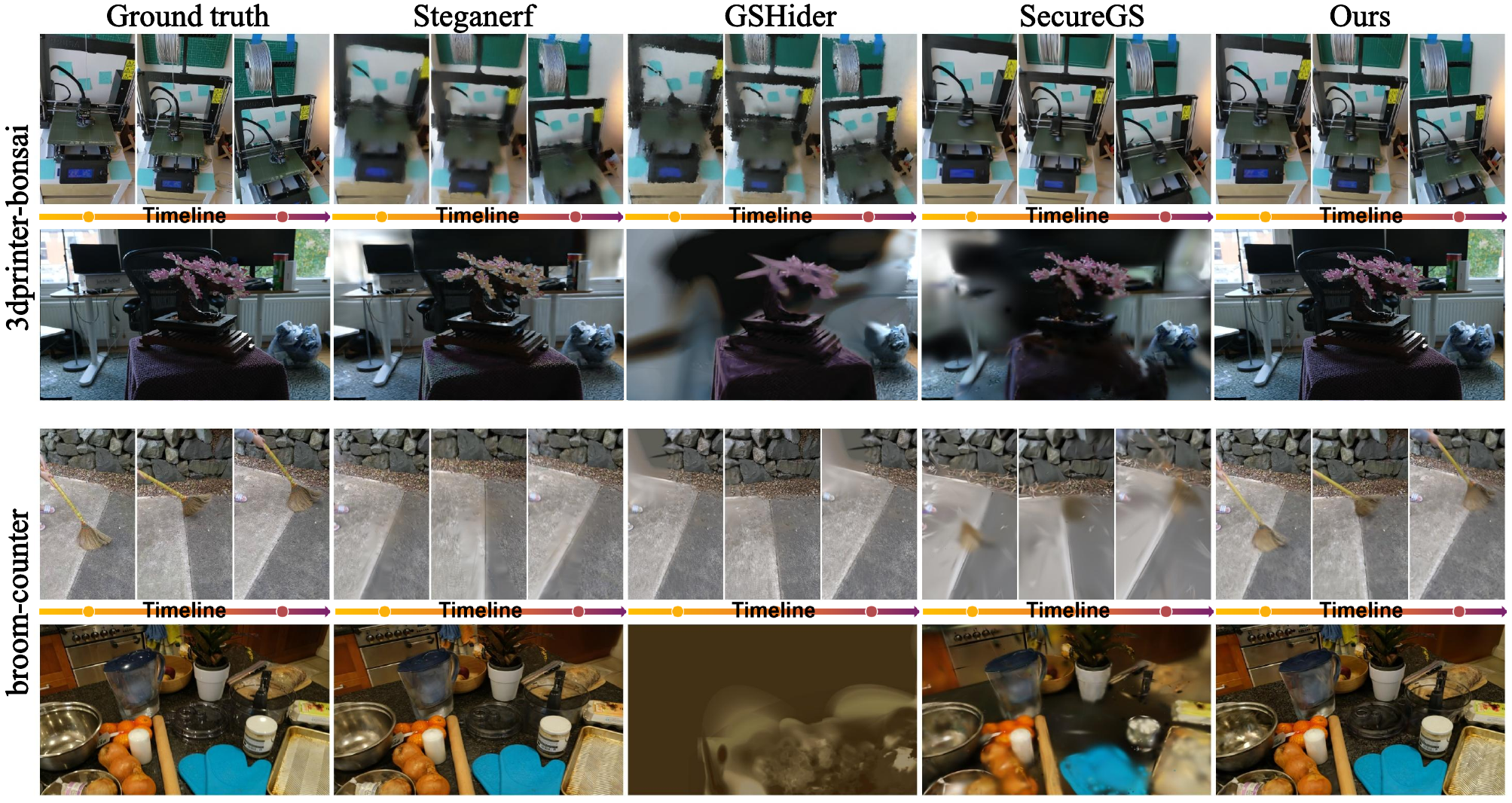}
  \caption{4D Steganography Visualization on HyperNeRF (4D Dynamic Scene) and Mip-NeRF 360 (3D Hidden Watermark Scene). For each scene, the top three images show rendering results of the dynamic scene at different time steps, while the bottom image shows the recovered static watermark scene. The results show that our method can accurately recover temporally varying 4D dynamic scenes while effectively ensuring the usability and visual fidelity of the embedded 3D hidden information.}
  \label{fig_compare_4d}
\end{figure*}
\subsection{Evaluation on 4D Steganography}

We evaluate the fidelity of 4D steganography by assessing the Novel View Synthesis (NVS) quality of both the original scenes and the hidden scenes. To ensure fair comparison, we extend all methods to 4D following 4DGS\cite{wu20234d}. Specifically, we augment each with a deformation field that warps canonical 3D Gaussians over time, using a two-stage training of static scene initialization followed by deformation field fine-tuning. Quantitative results are reported in Tab.~\ref{tab:4d_results}. As can be observed, our method consistently achieves the best performance across all NVS metrics on both scene types, demonstrating that the proposed framework effectively preserves the visual fidelity of the carrier scene while faithfully reconstructing the hidden scene.
Qualitative comparisons are presented in Fig.~\ref{fig_compare_4d}. Our method produces temporally coherent and visually appealing renderings with clean geometry and accurate appearance. In contrast, SecureGS suffers from noticeable rendering artifacts, especially in dynamic regions, while GS-Hider exhibits erroneous textures and color distortions that degrade overall visual quality. These results indicate that our \emph{Splats-in-Splats} formulation enables superior fidelity in 4D steganography, particularly under challenging dynamic and viewpoint-changing scenarios.

\subsection{Ablation Study}  
\subsubsection{Hash-Grid Guided Opacity Mapping} We demonstrate through ablation studies the critical role of this module in achieving complete and high-quality hidden information reconstruction. We compare our approach with two simple opacity mapping baselines:
(a) directly predicting opacity using an MLP, and
(b) employing an opacity mask combined with a CNN, as adopted in \cite{guo2024splats}.
Experimental results in Tab.~\ref{tab_ab_hashgrid} show that, with the inclusion of an opacity consistency loss to fuse the original and hidden scenes, these baseline methods fail to establish accurate geometric correspondences between the two scenes. In contrast, our Hash-Grid Guided Opacity Mapping consistently produces high-quality and geometrically coherent mappings.

\begin{table}[htb]
  \centering
  \small

    \setlength{\tabcolsep}{0.35mm}

    {
    \resizebox{0.85\linewidth}{!}{
    \begin{tabular}{c||ccc}
     \hline\thickhline
\rowcolor[HTML]{e6f0f0}Method&  $PSNR$ $\uparrow$&  $SSIM$  $\uparrow$& $LPIPS$  $\downarrow$\\
     \hline\hline
Naive MLP &  18.591 & 0.634 &  0.448 \\
Op-Mask+CNN &   20.876 & 0.704 & 0.384 \\
Hash-Grid Guided & 27.555 &0.828 & 0.265 \\
    \hline
    \end{tabular}%
}}

  \caption{Comparisons with image reconstruction baselines. }

  \label{tab_ab_hashgrid}%

\end{table}%

\begin{figure}[t]
  \centering
  \includegraphics[width=1\linewidth]{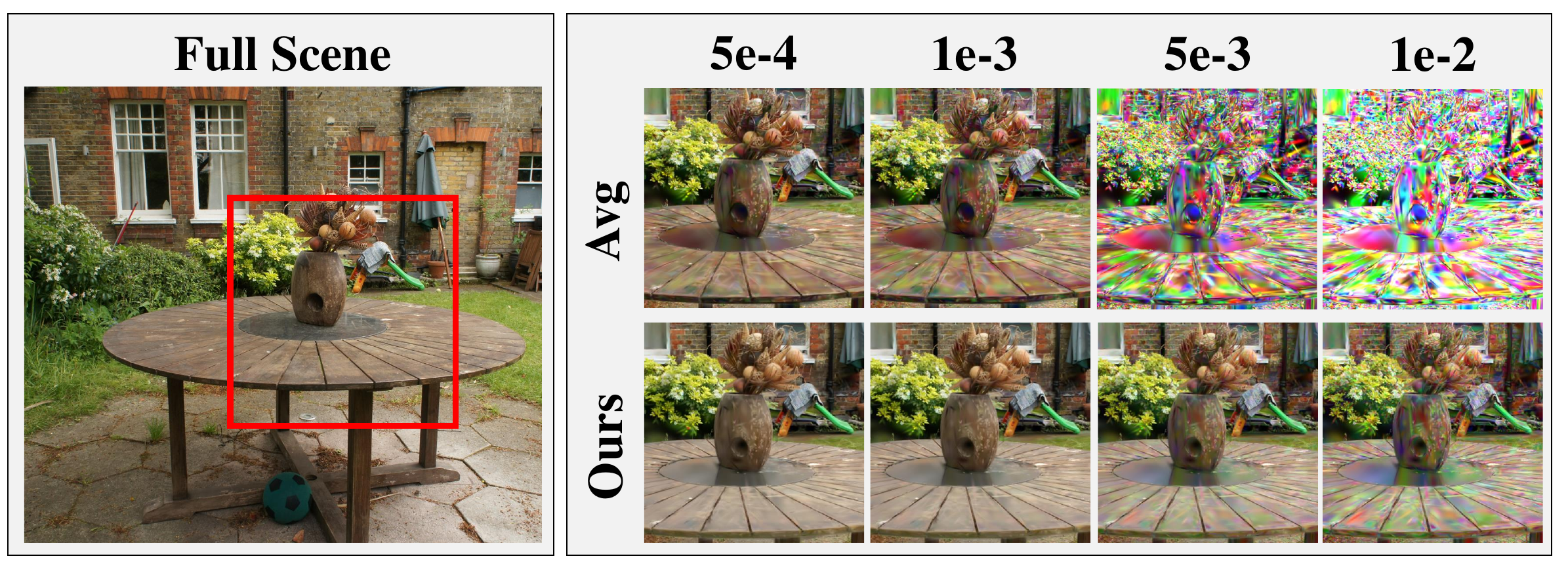}
  \caption{Visualization results under different noise levels. \textbf{Better viewed on screen with zoom in.}}
  \label{fig_noise}
\end{figure}

\subsubsection{Opacity consistency loss} We further investigate the impact of the opacity consistency loss on the coupling between the original and hidden scenes, as well as on the robustness of the proposed method, through ablation studies. As reported in Tab.~\ref{tab:pruning_attacks_summary}, we compare the performance of our method with and without the opacity consistency loss under various adversarial attacks. The results show that, when the opacity consistency loss is enabled, our method consistently achieves the best performance across all adversarial evaluations, significantly outperforming the baseline methods. In contrast, methods that disable the opacity consistency loss exhibit substantially degraded performance against many attack algorithms. These results demonstrate that the Opacity Gating mechanism inspired opacity consistency loss effectively enhances the robustness of our approach by enforcing surface-level coupling, thereby preventing the hidden information from being easily removed. More visualizations are provided in Fig.~\ref{fig:vis_robustness}.


\subsubsection{Importance-graded SH coefficient encryption} We further conduct ablations to illustrate why importance-graded SH coefficient encryption is necessary. We compare the performance of our method with the average encryption (AVG) under different levels of Gaussian noise, with noise levels ranging from 0.0005 to 0.01. Quantitative results in Tab.~\ref{tab_noise} demonstrate that importance-graded encryption offers a significant advantage in noise resistance, which enhances both security and robustness. Fig.~\ref{fig_noise} confirms that as the noise level increases, AVG encryption causes a dramatic degradation in the quality of the recovered hidden scene. In contrast, Splats in Splats++ exhibits strong robustness against high-level noise, ensuring successful extraction of the hidden 3D information.

\begin{table}[h]
  \centering
    \setlength{\tabcolsep}{1.3mm}

    {
    
    \begin{tabular}{cccccc}
     \hline\thickhline
\rowcolor[HTML]{e6f0f0}&\multicolumn{4}{c}{Gaussian noise level }& \\
 \cline{2-5}
\rowcolor[HTML]{e6f0f0}  \multirow{-2}{*}{Method}&0.0005 &0.001&0.005&0.01& \multirow{-2}{*}{Average} \\ 

     \hline
     AVG&24.167 &21.991 &11.442& 7.471& 16.267 \\
     Ours&\textbf{24.577} &\textbf{24.509} &\textbf{22.797}& \textbf{20.032}& \textbf{22.979}\\
   \hline
    \end{tabular}%
}
  \caption{Quantitative results of PSNR for rendered hidden images under different noise levels.}
  \label{tab_noise}%
\end{table}


\subsection{Further Applications}
\subsubsection{Image Embedding} 
\begin{figure}[t]

  \centering
  \includegraphics[width=1\linewidth]{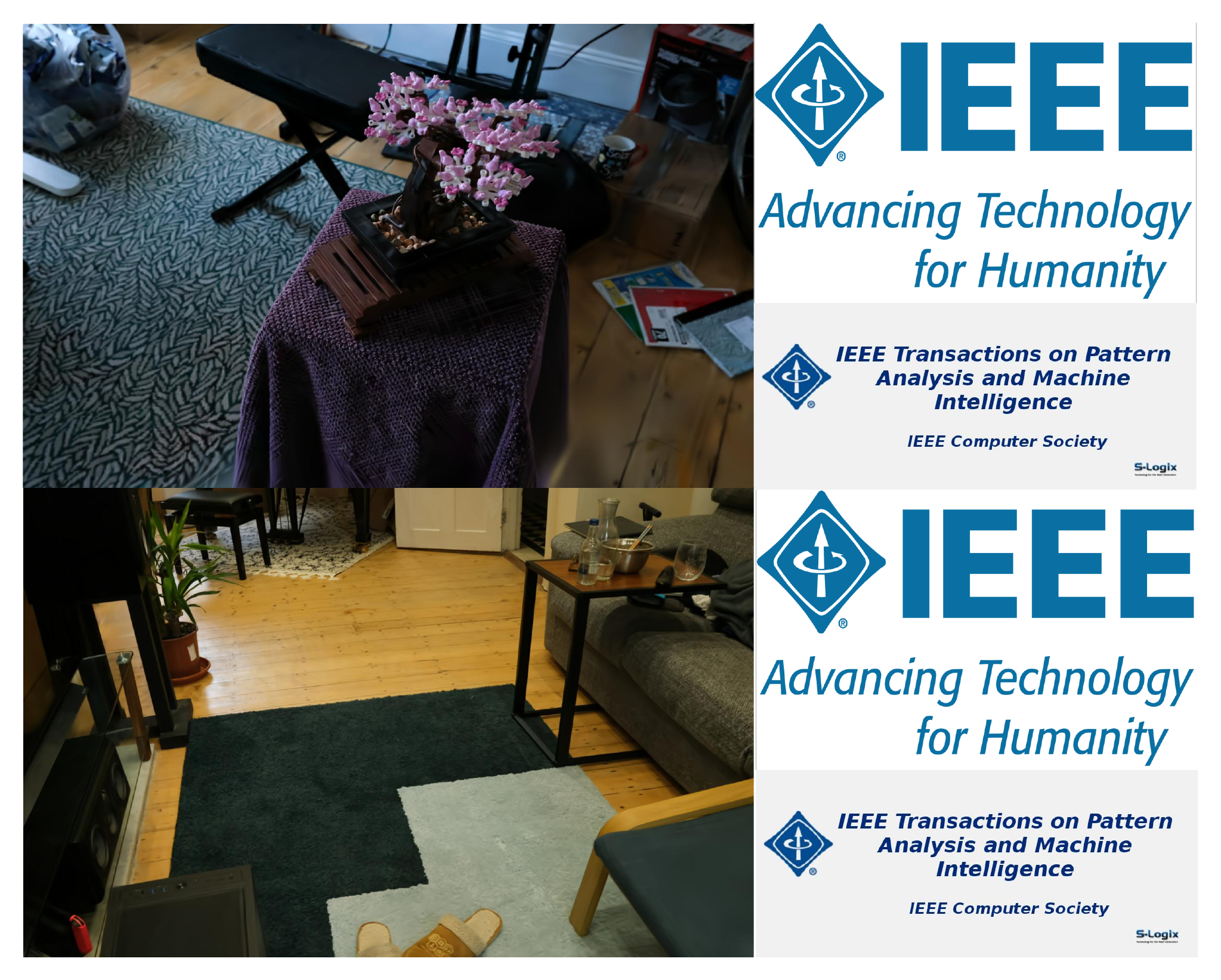}

  \caption{Visualization results of image embedding. We embed IEEE logo (right) into 3D scenes (left). }
  \label{fig_imghide}

\end{figure}

Our method can also embed images into the original 3D scene. In fact, this represents a degenerate case of embedding 3D scenes. Similar to GS-Hider, we treat the image embedding task as a scene embedding task with only a single viewpoint. The visualization results in Fig.~\ref{fig_imghide} demonstrate that Splats in Splats++ can embed and recover high-quality hidden images from the original scenes with minimal impact on the original scene itself.
\subsubsection{Downstream tasks}
\begin{figure}[t]

  \centering
  \includegraphics[width=1\linewidth]{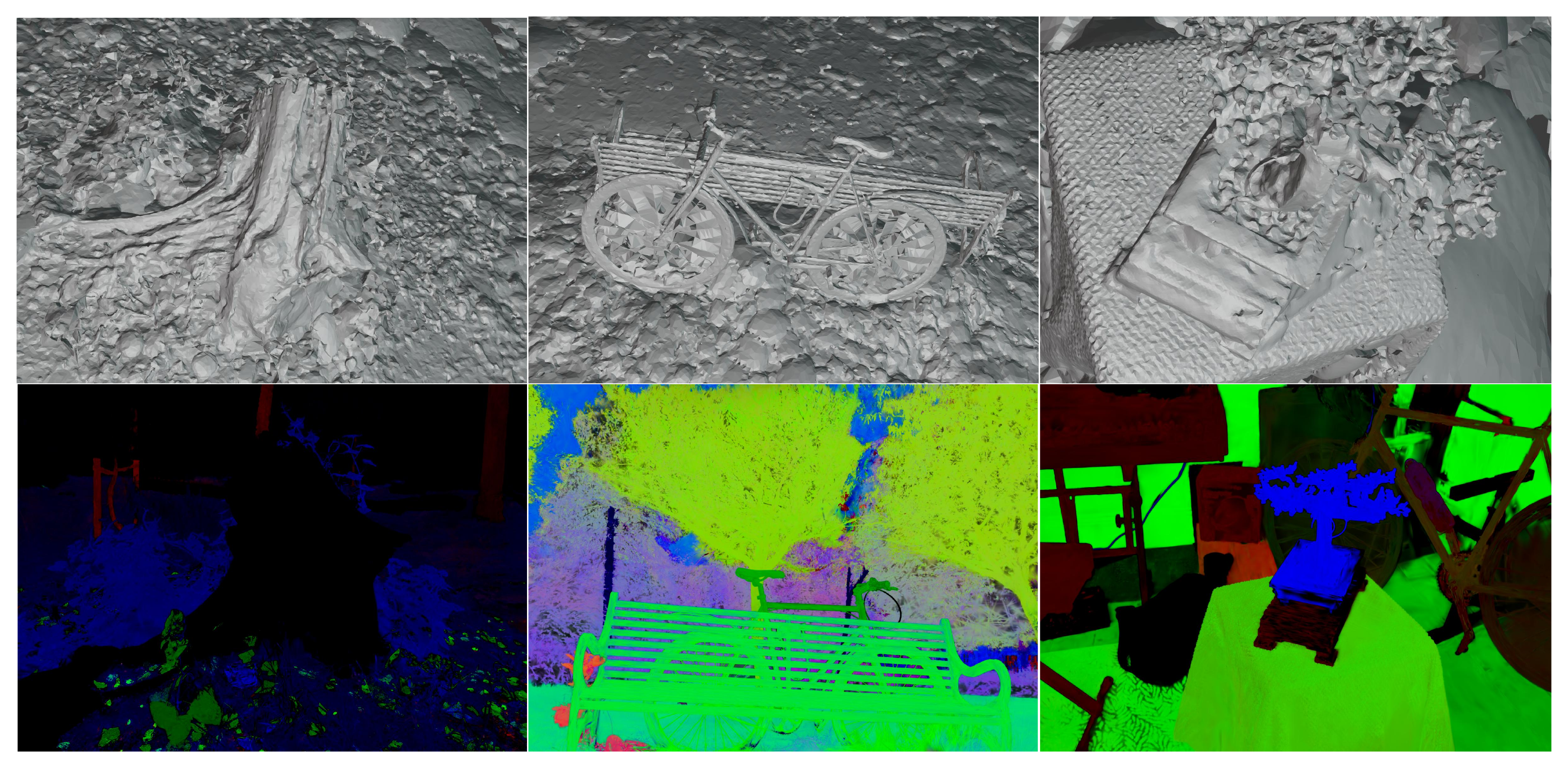}
  \caption{Visualization results of downstream tasks. Our encrypted/decrypted 3DGS scene can be directly used for: \textbf{Top:} Mesh extraction with Sugar (CVPR 24). \textbf{Bottom:} 3D segmentation with Langsplats  (CVPR 24 Highlight).}
  \label{fig_downstream_app}
\end{figure}

Our method provides an excellent user experience, demonstrated by its compatibility with all 3DGS rendering pipelines and engines, as well as its seamless integration into downstream tasks utilizing 3DGS.  Fig.~\ref{fig_downstream_app} demonstrates that both original and hidden scenes of our method can be directly applied to mesh extraction \cite{guedon2024sugar} and 3D segmentation \cite{qin2024langsplat} with exciting performance. 


\section{Conclusion}
We propose Splats in Splats++, a unified steganography framework that embeds 3D content directly into vanilla 3DGS representations without modifying their attributes or rendering pipeline. By operating entirely within the native 3DGS formulation, our method preserves rendering fidelity, real-time performance, and full compatibility with standard rendering engines and downstream tasks, addressing the usability limitations of existing 3DGS steganography approaches. Through importance-graded modulation of SH coefficients, implicit opacity mapping, and density-gradient driven geometric coupling, Splats in Splats++ achieves imperceptible embedding while maintaining strong robustness against structural perturbations and specific adversarial attacks designed for 3DGS. Extensive experiments across diverse datasets and scenarios demonstrate consistent improvements in visual fidelity, efficiency, and robustness over prior methods. Moreover, the proposed framework generalizes seamlessly to advanced settings such as 4DGS, highlighting its extensibility and practical value for real-world 3D asset protection and attribution.

\textbf{Limitations:} In our method, embedding hidden information involves modifying the higher-order Spherical Harmonics of the original scene, which can affect the rendering quality of specular or highly reflective surfaces. Consequently, applying this framework to certain specialized 3D assets, such as mirror-like or highly glossy objects, may pose challenges for high-fidelity reconstruction, and view-dependent effects could be slightly compromised. In addition, because our method is specifically designed for 3D steganography, it does not allow straightforward modification to embed arbitrary bitstreams, which prevents direct comparison with some mainstream 3D watermarking techniques. For future work, we plan to explore more general and consistent 3D ownership protection frameworks that can accommodate a wider range of 3D assets and applications, ultimately contributing to a more robust ecosystem for 3D content creation and distribution.



{
\bibliographystyle{IEEEtran}
\bibliography{ref}
}

\begin{IEEEbiography}[{\includegraphics[width=1in,height=1.25in,clip,keepaspectratio]{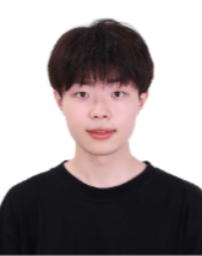}}]{Yijia~Guo}
  received the B.S. degree from Beihang University, Beijing, China, in 2022. He is currently working toward the Ph.D. degree with the National Engineering Research Center of Video Technology, School of Computer Science, Peking University. His research interests are focused on 3DV.
\end{IEEEbiography}
\vspace{-5mm}
\begin{IEEEbiography}[{\includegraphics[width=1in,height=1.25in,clip,keepaspectratio]{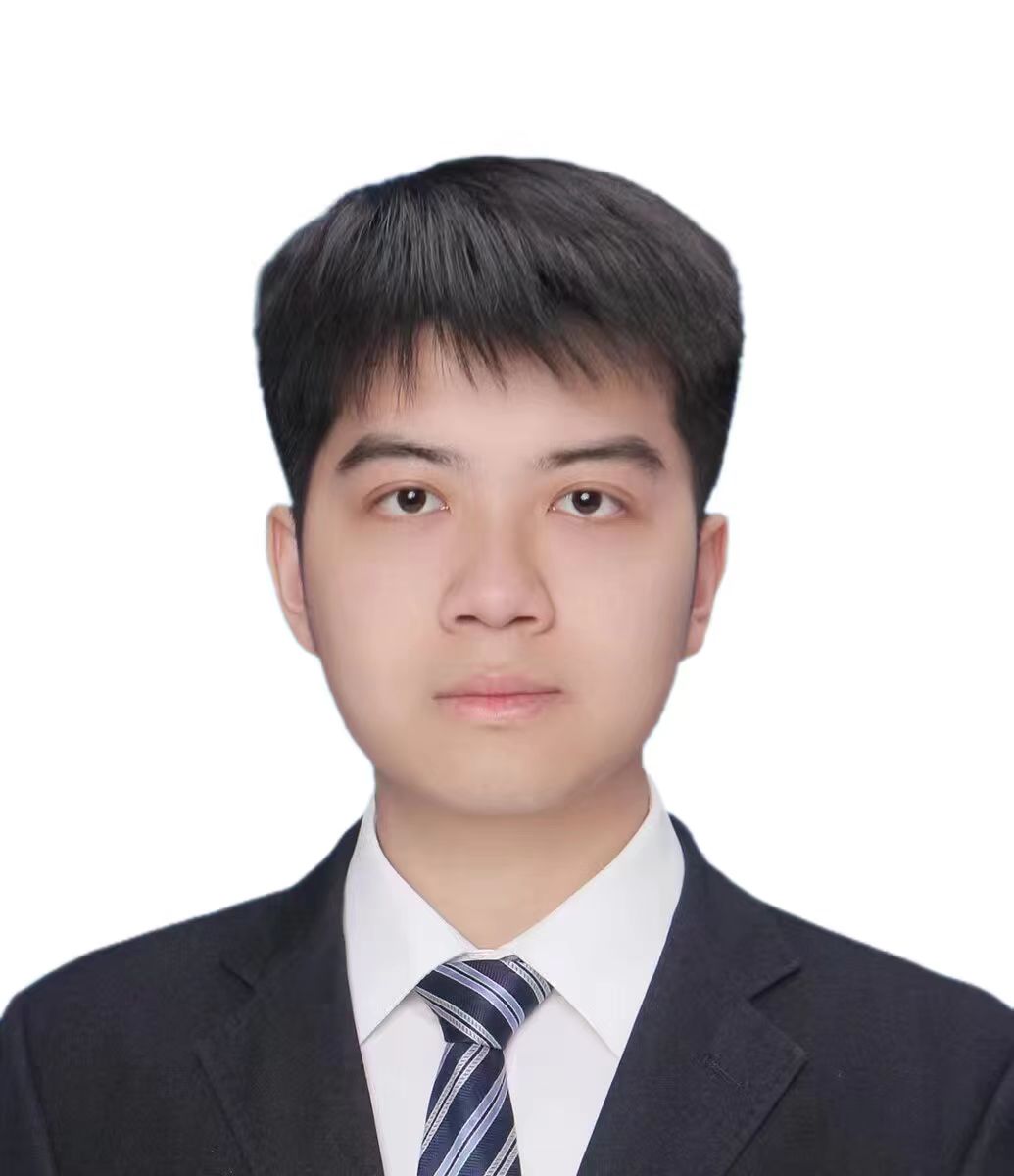}}]{Wenkai~Huang}
  received the B.S. degree in Information Security (IEEE Honors Class) from Shanghai Jiao Tong University, Shanghai, China, in 2023. He is currently pursing the Ph.D degree in School of Computer Science, Shanghai Jiao Tong University, Shanghai, China. His research interests include backdoor learning, federated learning and data security.
\end{IEEEbiography}
\vspace{-5mm}

\begin{IEEEbiography}[{\includegraphics[width=1in,height=1.25in,clip,keepaspectratio]{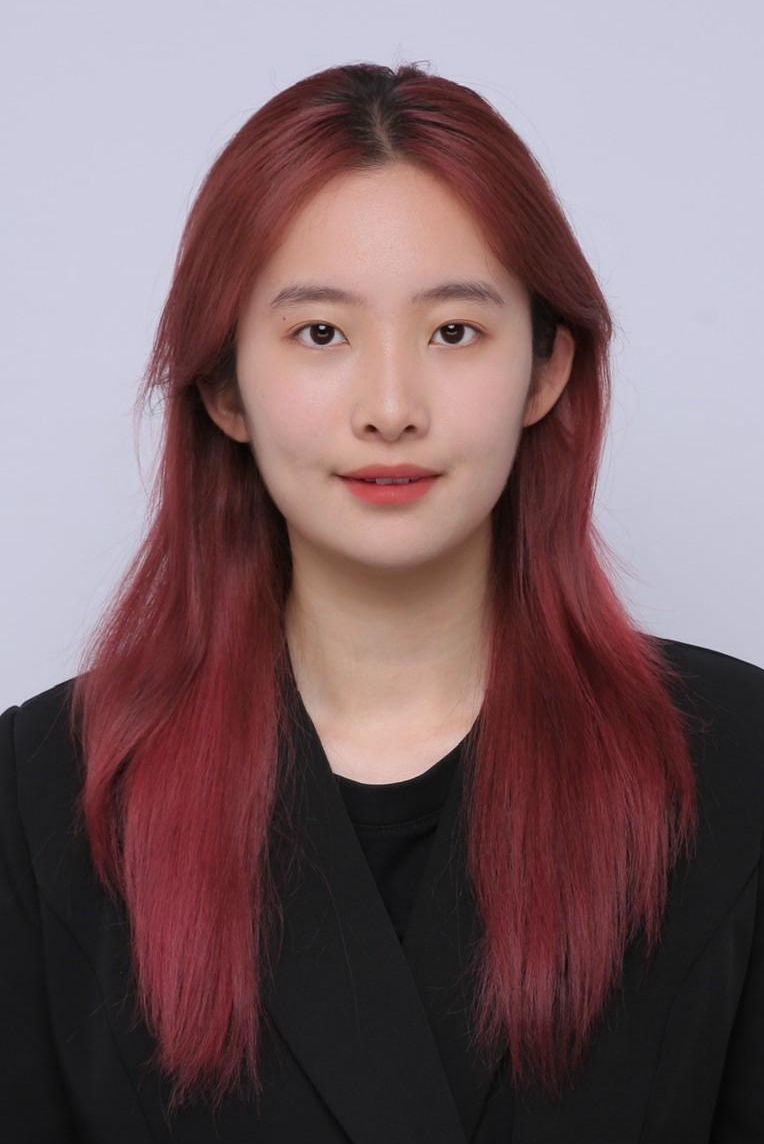}}]{Tong~Hu}
received the B.S. degree in software engineering from Shanghai Jiao Tong University, Shanghai, China, in 2025. She is currently working toward the M.S. degree with the School of Software and Microelectronics, Peking University, Beijing, China. Her research interests are focused on 3DV.
\end{IEEEbiography}
\vspace{-5mm}

\begin{IEEEbiography}[{\includegraphics[width=1in,height=1.25in,clip,keepaspectratio]{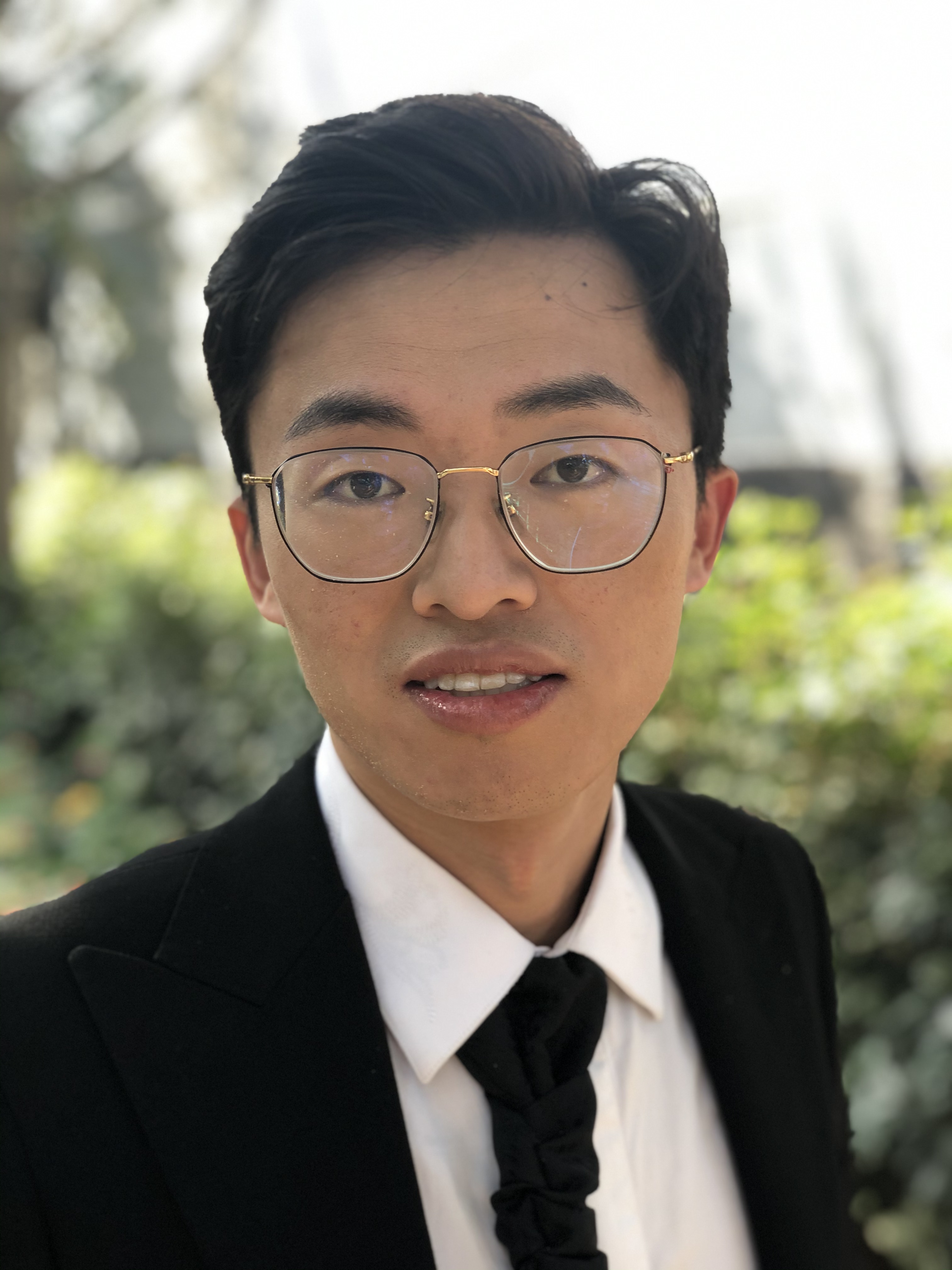}}]{Gaolei~Li}
(Member, IEEE) is currently an Associate Professor with the School of Computer Science, Shanghai Jiao Tong University, Shanghai, China. He has published more than 100 papers, including IEEE TIFS, IEEE TDSC, NDSS, ACM CCS, ACM MM, and AAAI. His research interests include machine learning security and privacy-preserving. He has obtained many awards, including the Outstanding Paper Award of IEEE ISPA 2024, the Third Prize of Science and Technology Progress Award of China Electric Power Development Acceleration Committee in 2024, the Best Conference Paper Award of China Cryptology Society in 2020, and the Best Conference Paper Award of IEEE CSIM Committee in 2018. He also served as a TPC member for CVPR 2024–2025, AAAI 2023\&2024\&2025, ACM MM 2023\&2024\&2025, and ICLR 2025. Besides, he also serves a reviewer for IEEE TIFS, IEEE TDSC, IEEE TMC, IEEE JSAC, IEEE/ACM ToN, IEEE TSC, IEEE TNNLS, and IEEE TGCN.
\end{IEEEbiography}
\vspace{-5mm}

\begin{IEEEbiography}
[{\includegraphics[width=1in,height=1.25in,clip,keepaspectratio]{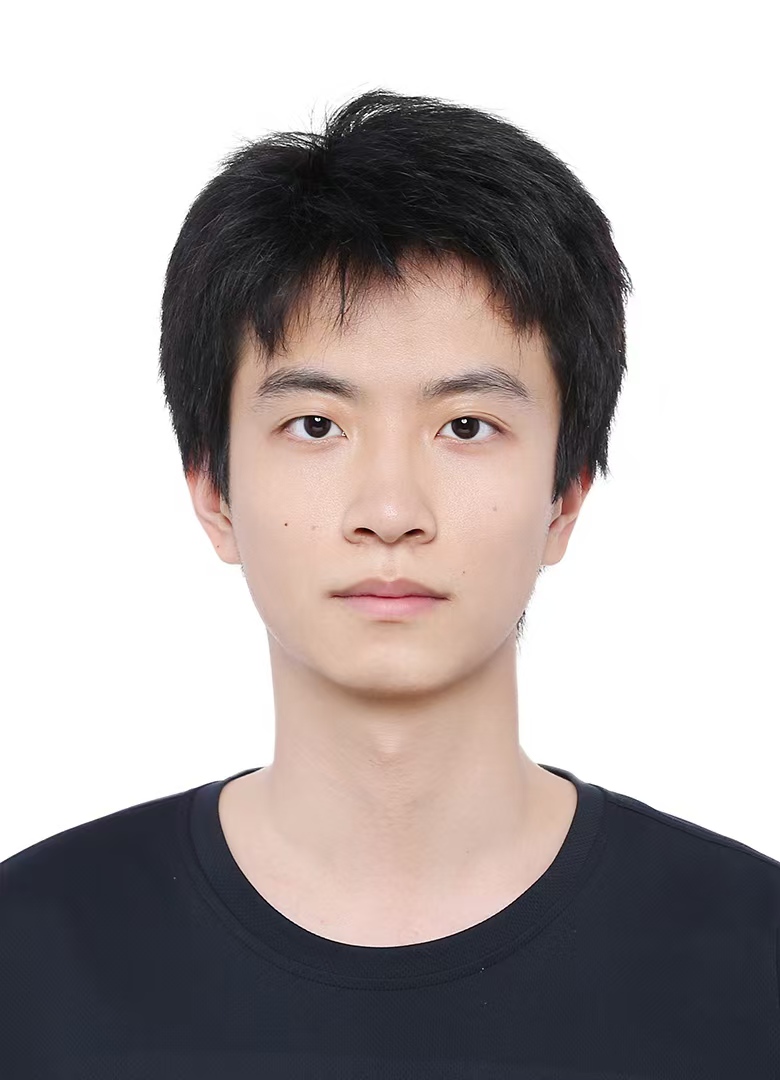}}]{Yang~Li}
 received the B.S. degree from the School of 
Electrical Engineering and Computer Science, Peking University, Beijing, China, in 2025. He is currently working toward the M.S degree with the School of Software and Microelectronics, Peking University. His research interests are focous on Computer Vision.
\end{IEEEbiography}
\vspace{-5mm}

\begin{IEEEbiography}[{\includegraphics[width=1in,height=1.25in,clip,keepaspectratio]{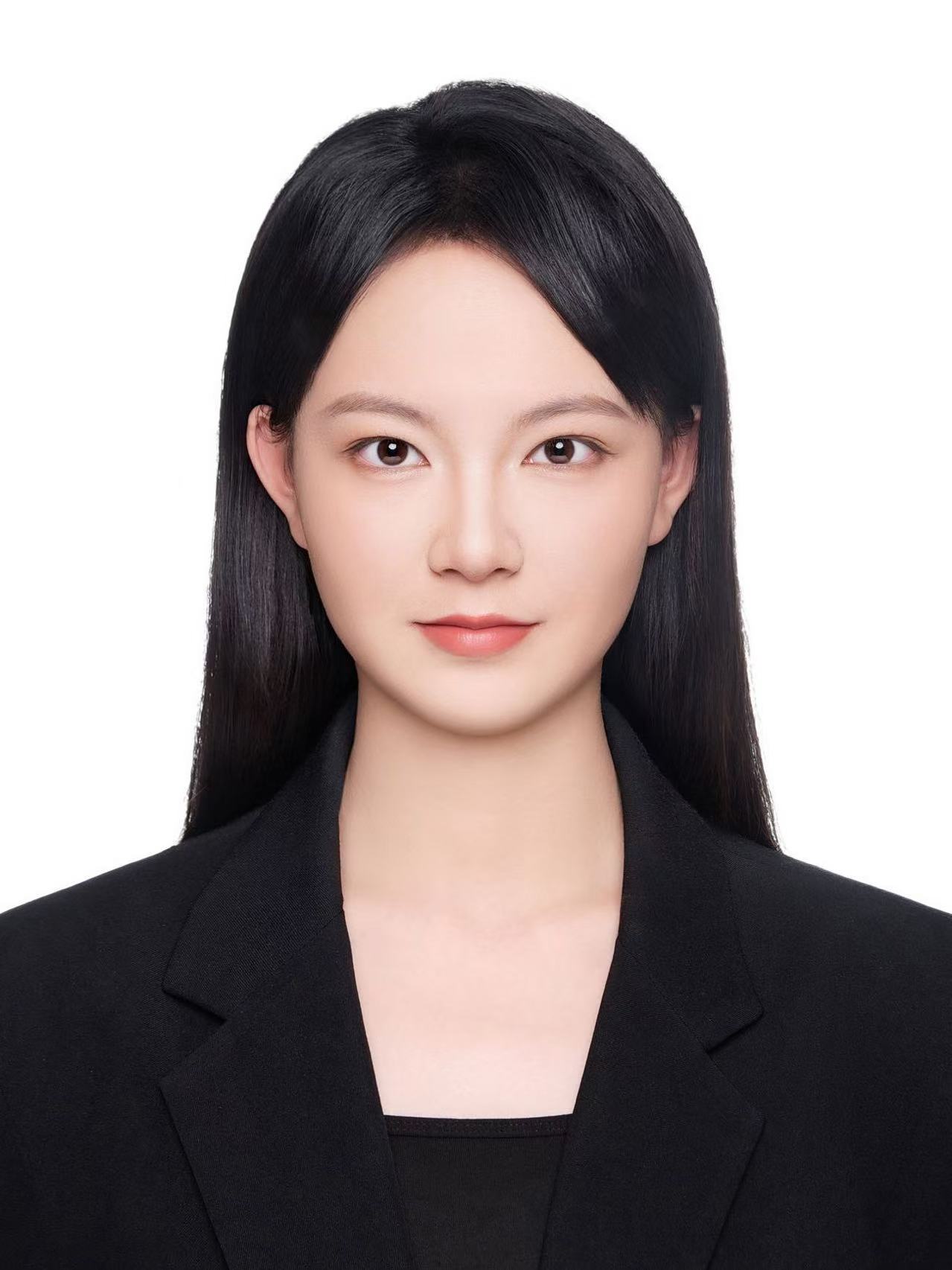}}]{Yuxin~Hong}
  received the B.E. degree in software engineering from Changchun University, Changchun, China, in 2025. She is currently pursuing the M.Eng. degree in software engineering with Nanchang University, Nanchang, China. Her research interests are focused on 3DV.
\end{IEEEbiography}
\vspace{-5mm}
\begin{IEEEbiography}
[{\includegraphics[width=1in,height=1.25in,clip,keepaspectratio]{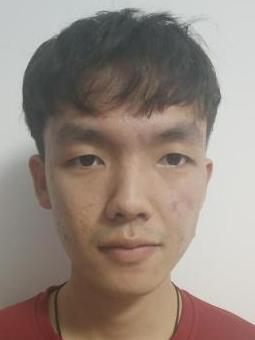}}]{Liwen~Hu}
received the B.S. degree from University
of Electronic Science and Technology
of China, Chengdu, China, in 2019 and the Ph.D. degree with
the National Engineering Research Center of
Video Technology, School of Computer Science,
Peking University in 2025.  Currently, he works as a postdoctor at Peking University. His research interests are focused
on application of neuromorphic cameras.
\end{IEEEbiography}
\vspace{-5mm}
\begin{IEEEbiography}[{\includegraphics[width=1in,height=1.25in,clip,keepaspectratio]{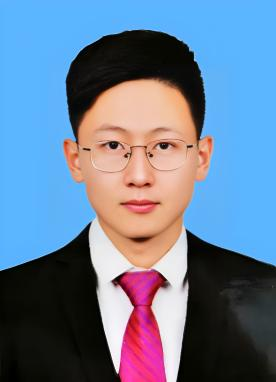}}]{Xitong~Ling}
received the B.S. degree in aircraft propulsion engineeringfrom the Beihang University, China, in 2023. He is currently pursuing the M.E. degree in Tsinghua University, Beijing. His research interests include multiple instance learning for whole slide image analysis, multimodal learning for medical foundation model and 3D-Vision. He has published academic papers in international computer conferences and Journal of Biomedical Sciences.
\end{IEEEbiography}
\vspace{-5mm}
\begin{IEEEbiography}[{\includegraphics[width=1in,height=1.25in,clip,keepaspectratio]{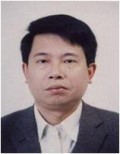}}]{Jianhua~Li} (Senior Member, IEEE)
is a professor/Ph.D. supervisor and the dean of Institute of Cyber Science and Technology, Shanghai Jiao Tong University, Shanghai, China. He is also the director of National Engineering Laboratory for Information Content Analysis Technology, the director of Engineering Research Center for Network Information Security Management and Service of Chinese Ministry of Education. He has published more than 400 papers, including IEEE TDSC, IEEE TIFS, IEEE TMC, IEEE TETC, ACM CCS, ACM MobiHoc, and IEEE Infocom. He got many awards, including the Second Prize of National Technology Progress Award of China in 2005, first Prize of Science and Technology Progress Award of the Ministry of Education, first Prize of Science and Technology Progress Award of Shanghai, Best paper award of IEEE CSIM committee in 2016, IEEE TETC in 2017, ESI Highly Cited Scientist from 2022 to 2024. His research interests include information security, signal process, computer network communication, etc.
\end{IEEEbiography}
\vspace{-5mm}

\begin{IEEEbiography}[{\includegraphics[width=1in,height=1.25in,clip,keepaspectratio]{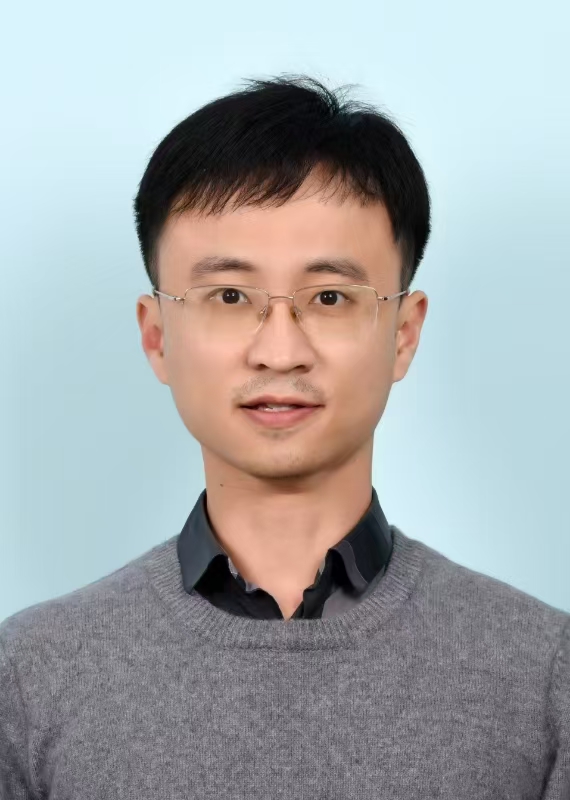}}]{Shengbo~Chen}
 received the B.E. and M.E. degrees from the Department of Electronic Engineering, Tsinghua University, Beijing, China, in 2006 and 2008, respectively, and the Ph.D. degree from Ohio State University, Columbus, OH, USA, in 2013. From 2013 to 2019, he was a Senior Research Engineer with the Qualcomm Research Center, San Diego, CA, USA. He is currently a professor with the School of Software, Nanchang University, China. He holds more than 50 U.S. patents on 5G and AI area. Dr. Chen received the Best
Student Paper Award for WiOpt 2013.
\end{IEEEbiography}
\vspace{-70mm}

\begin{IEEEbiography}[{\includegraphics[width=1in,height=1.25in,clip,keepaspectratio]{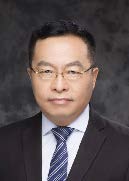}}]{Tiejun~Huang}
   (Senior Member, IEEE) received the B.Sc. and M.Sc. degrees in computer science from Wuhan University of Technology, China in 1992 and 1995, respectively, and the Ph.D. degree in pattern recognition and image analysis from Huazhong (Central China) University of Science and Technology in 1998. He is currently a professor with the School of Computer Science, Peking University, and the Director of the Beijing Academy for Artificial Intelligence. His research areas include visual information processing and neuromorphic computing. He is a Fellow of CAAI, CCF, CSIG and vice chair of the China National General Group on AI Standardization. He published 300+ peer-reviewed papers on leading journals and conferences, and also co-editor of 4 ISO/IEC standards, 5 National standards and 4 IEEE standards. He holds 100+ granted patents. Professor Huang received National Award for Science and Technology of China (Tier-2) for three times (2010, 2012, 2017).
\end{IEEEbiography}
\vspace{-70mm}
\begin{IEEEbiography}[{\includegraphics[width=1in,height=1.25in,clip,keepaspectratio]{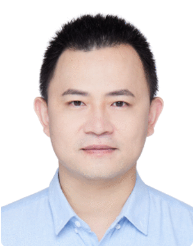}}]{Lei~Ma}
(Member, IEEE) obtained the BS degree from Zhejiang University, the MS degree from Digital ART (Augmented Reality Tech) Laboratory of Shanghai Jiao Tong University, and the PhD degree from the State Key Laboratory of Computer Science, Chinese Academy of Sciences. He is an associate research professor with Peking University and the department head of the Life Simulation Research Center, Beijing Academy of Artificial Intelligence (BAAI). During 2010-2012, he worked for Autodesk China Research and Development Center as a graphic engineer. He was also a co-founder of an AR start-up. His research interests include realistic image synthesis, geometry processing, rendering of complex scenes, virtual reality, and artificial intelligence.
\end{IEEEbiography}
\vspace{-5mm}

\end{document}